\documentclass{article} 
\usepackage{iclr2026_conference,times}

\usepackage{amsmath,amsfonts,bm}

\def\eqref#1{equation~\ref{#1}}

\def\1{\bm{1}}

\DeclareMathAlphabet{\mathsfit}{\encodingdefault}{\sfdefault}{m}{sl}
\SetMathAlphabet{\mathsfit}{bold}{\encodingdefault}{\sfdefault}{bx}{n}

\usepackage{hyperref}
\usepackage{url}
\usepackage{graphicx}
\usepackage{caption}

\usepackage{algorithm}
\usepackage{algorithmic}
\usepackage{booktabs}
\usepackage{float}
\usepackage{amsmath}
\usepackage{multirow}
\usepackage{multicol}
\usepackage{array}

\usepackage{newfloat}
\usepackage{listings}

\usepackage{xcolor}

\lstdefinestyle{promptstyle}{
    basicstyle=\ttfamily\small\linespread{0.9},
    backgroundcolor=\color{gray!5}, 
    frame=single, 
    framerule=0.8pt, 
    rulecolor=\color{gray!60}, 
    framesep=6pt, 
    xleftmargin=8pt, 
    xrightmargin=8pt, 
    breakindent=0pt, 
    breakautoindent=false,
    breaklines=true, 
    breakatwhitespace=true, 
    postbreak={}, 
    aboveskip=12pt, 
    belowskip=12pt, 
    tabsize=2,
    showstringspaces=false,
    numbers=none, 
    captionpos=t, 
    escapeinside={@}{@}, 
    prebreak=\mbox{}, 
    postbreak=\mbox{} 
}

\title{Lego-Edit: A General Image Editing Framework with Model-Level Bricks and MLLM Builder}

\author{\textbf{Qifei Jia$^*$, Yu Liu$^*$, Yajie Chai, Xintong Yao, Qiming Lu, Yasen Zhang$^\dagger$,} \\ \textbf{Runyu Shi, Ying Huang, Guoquan Zhang} \\
Xiaomi Corporation\\
Beijing, China
}

\iclrfinalcopy 
\begin{document}

\def\thefootnote{$^{*}$}\footnotetext{Equal contribution}
\def\thefootnote{$^{\dagger}$}\footnotetext{Corresponding author}
\def\thefootnote{\arabic{footnote}}

\maketitle

\begin{figure}[!htb]
\centering
  \includegraphics[width=\linewidth]{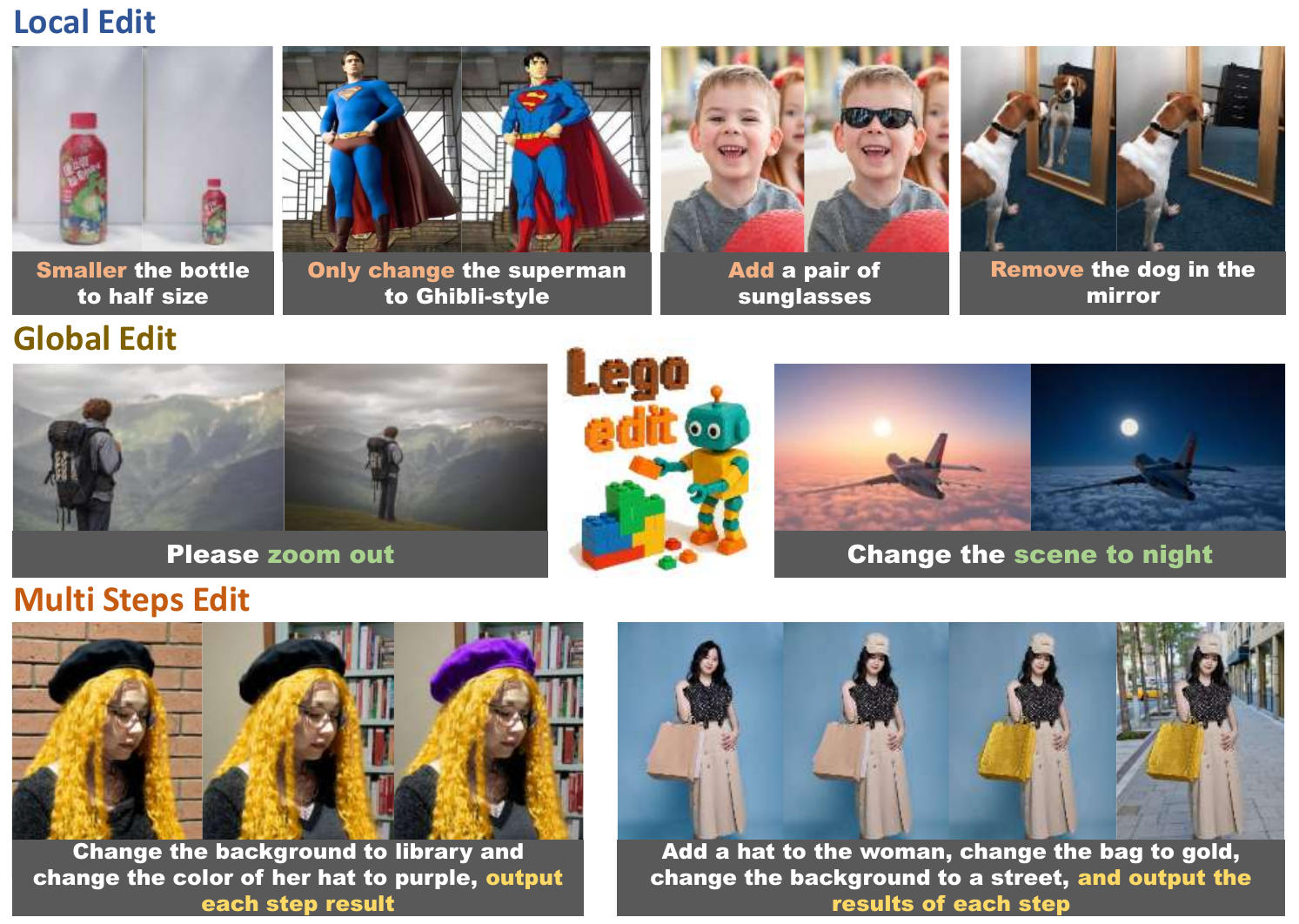}
  \caption{Comparison of end-to-end approach, API-based agent with curated workflows, and our method.}
  \label{fig:intro}
\end{figure}

\begin{abstract}
Instruction-based image editing has garnered significant attention due to its direct interaction with users. However, real-world user instructions are immensely diverse, and existing methods often fail to generalize effectively to instructions outside their training domain, limiting their practical application. To address this, we propose Lego-Edit, which leverages the generalization capability of Multi-modal Large Language Model (MLLM) to organize a suite of model-level editing tools to tackle this challenge. Lego-Edit incorporates two key designs: (1) a model-level toolkit comprising diverse models efficiently trained on limited data and several image manipulation functions, enabling fine-grained composition of editing actions by the MLLM; and (2) a three-stage progressive reinforcement learning approach that uses feedback on unannotated, open-domain instructions to train the MLLM, equipping it with generalized reasoning capabilities for handling real-world instructions. Experiments demonstrate that Lego-Edit achieves state-of-the-art performance on GEdit-Bench and ImgBench. It exhibits robust reasoning capabilities for open-domain instructions and can utilize newly introduced editing tools without additional fine-tuning. 

Code is available: https://github.com/xiaomi-research/lego-edit.
\end{abstract}

\section{Introduction}

Instruction-based image editing methods accept natural language instructions as input and modify the input image accordingly. It enables intuitive human-computer interaction through natural language, offering broad application potential. However, the significant diversity inherent in real-world editing instructions poses a substantial challenge for image editing systems in handling flexible user commands.

Existing approaches for instruction-based editing are broadly classified into two categories. End-to-end methods \cite{flux2024,liu2025step1x} train a single generative model to learn both instruction comprehension and pixel mapping for editing implicitly. As shown in Fig.~\ref{fig:intro}~(a), these methods are primarily constrained by the fixed instruction patterns within their training data, and consequently struggle to generalize well even with massive training datasets.

In contrast, agent-based schemes utilize MLLMs to explicitly interpret editing instructions and invoke editing tools to execute the requested modifications. Prior research \cite{xue2025comfybench} often relies on curated prompts to guide the agents, but such prompt-driven approaches lack a deep understanding of editing tools, impeding agents' ability to organize them effectively. Subsequent studies \cite{guo2025comfymind} attempt to alleviate this burden by constructing complex predefined workflows as task-level tools for agents to invoke, as shown in Fig.~\ref{fig:intro}~(b). However, this strategy inherently limits the framework's capacity to handle instructions that deviate from the predefined workflows.

To address the challenge of processing flexible real-world instructions for image editing, we propose a novel framework, named Lego-Edit. It employs a fine-tuned MLLM as an agent, termed Builder. The Builder leverages its enhanced reasoning capability to organize a set of specialized pre-trained editing models, called Bricks, enabling precise execution of diverse editing instructions, as shown in Fig.~\ref{fig:intro}~(c). It incorporates two key design innovations:

\begin{figure}[!h]
\centering
\includegraphics[width=0.7\linewidth]{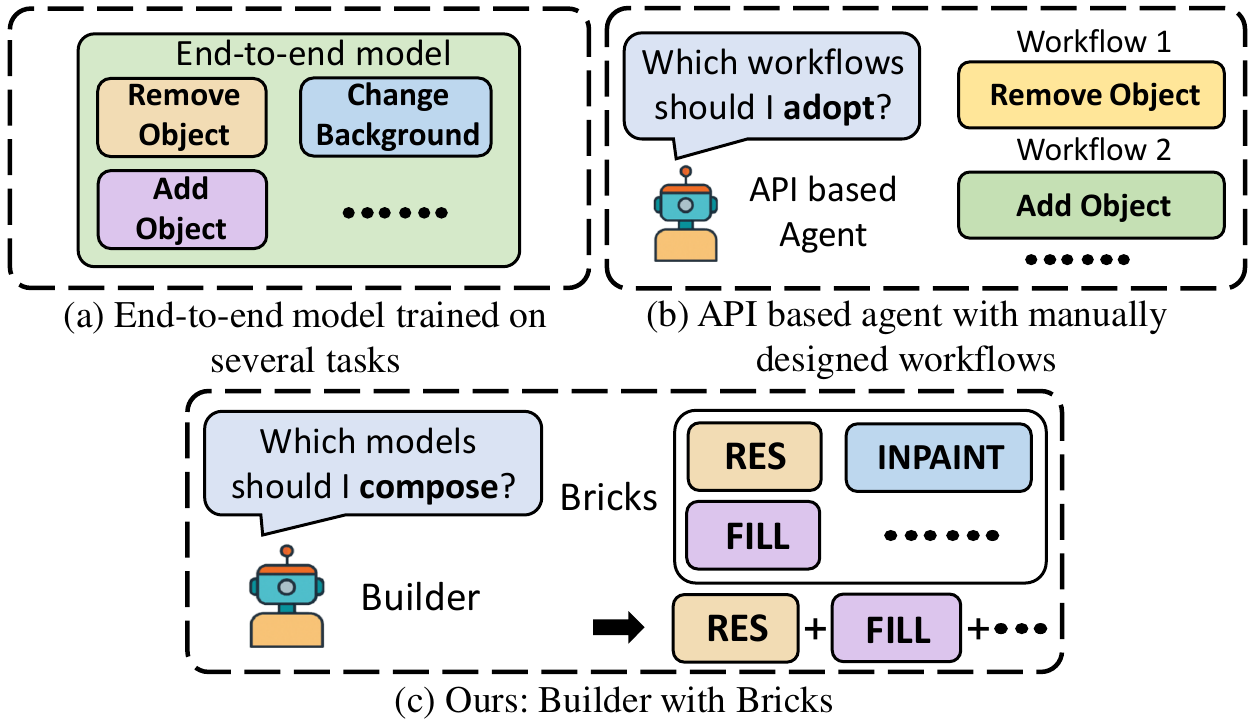}
\caption{Comparison of end-to-end approach, API-based agent with curated workflows, and our method.}
\label{fig:intro}
\end{figure}

\textbf{Model-Level Tools:} We train and integrate a suite of specialized models as editing tools. This model-level design provides the Builder with high flexibility for composition and enables individual tools to achieve superior performance for their specific functions with less training data. 

\textbf{Three-Stage Progressive Reinforcement Learning Training Strategy:} We first employ a Supervised Fine-Tuning (SFT) stage, followed by a Reinforcement Learning (RL) stage to train our Builder on specific editing tasks with the ground truth, establishing fundamental reasoning capabilities and tool usage knowledge. Subsequently, we conduct an additional RL stage utilizing abundant unlabeled instructions beyond the specific tasks, where a large-scale critic model provides feedback. This process enhances our Builder's capability to handle flexible instructions.

Benefiting from these designs, our Lego-Edit reliably reasons about and executes flexible editing instructions and can integrate novel tools without additional training. Furthermore, it achieves state-of-the-art results on the GEdit-Bench and ImgBench benchmarks.
 
In summary, our principal contributions are threefold:
\begin{itemize}
    \item We propose Lego-Edit, an instruction-based image editing framework that utilizes a reinforcement learning fine-tuned MLLM agent to coordinate model-level editing tools for executing flexible real-world instructions.

    \item We introduce a three-stage progressive reinforcement learning training strategy that provides feedback using unlabeled data, significantly enhancing the reasoning and tool composition capabilities of MLLM.

    \item Extensive experiments demonstrate that Lego-Edit achieves SOTA performance on GEdit-Bench and ImgBench. The framework also exhibits strong generalization in processing flexible open-domain instructions and could integrate new tools without retraining.
\end{itemize}

\section{Related Works}

\noindent\textbf{Instruction-based Image Editing:} Instruction-driven image editing, which emerged from InstructPix2Pix \cite{brooks2023instructpix2pix}, provides a natural user interface for image editing. The majority of existing methods rely on diffusion models \cite{gao2025eraseanything, han2024ace, hertz2022prompt}; some methods incorporate Multi-modal Large Language Models (MLLMs) to achieve more precise edits. For instance, SmartEdit leverages embeddings derived from nuanced edit semantics by MLLMs \cite{huang2024smartedit}.

Models trained on massive datasets (e.g., FLUX \cite{flux2024}, HiDream \cite{cai2025hidream}, Step1X \cite{liu2025step1x}) exhibit strong performance across various editing tasks. However, generalizing effectively across flexible editing instructions remains a critical challenge for them. Recently developed unified visual understanding and generation models (e.g., ILLUME++ \cite{huang2025illume+}, GPT-4o \cite{hurst2024gpt}, Bagel \cite{deng2025emerging}, UniWorld \cite{lin2025uniworld}), trained on broader datasets encompassing multiple tasks like image captioning and image editing, demonstrate enhanced generalization capabilities yet are still constrained.

In contrast, our proposed approach achieves superior instruction generalization with minimal training data by orchestrating specialized editing tools via a fine-tuned agent.

\noindent\textbf{MLLMs as Agents:} Autonomous agents capable of utilizing tool calls have garnered significant research interest. Many approaches primarily leverage LLMs to invoke tools \cite{qin2023toolllm, du2024anytool, zheng2024toolrerank}. With advances in MLLMs, considerable effort has focused on employing or fine-tuning MLLMs for multi-modal agent applications. For instance, VisualToolAgent \cite{huang2025visualtoolagent} aligns models with tools via reinforcement learning, and SeeAct \cite{zheng2024gpt} integrates MLLMs with grounding capabilities for WebUI interactions.

Within the domains of image generation and editing, research explores MLLMs as agents for tool invocation. ComfyAgent \cite{xue2025comfybench} adapts prompts to call ComfyUI tools via code execution, but this code-based invocation constrains its performance. ComfyMind \cite{guo2025comfymind} manually defines multiple pipelines for agent-driven tool use, ensuring high success rates but limiting operational flexibility.

Notably, our approach employs reinforcement learning to equip the MLLM with compositional tool-usage knowledge and reasoning ability. Combined with model-level tools, this framework achieves high success rates, superior editing performance, and robust generalization to diverse instructions.

\section{Method}
\label{sec:method}

\begin{figure*}
\centering
\includegraphics[width=\linewidth]{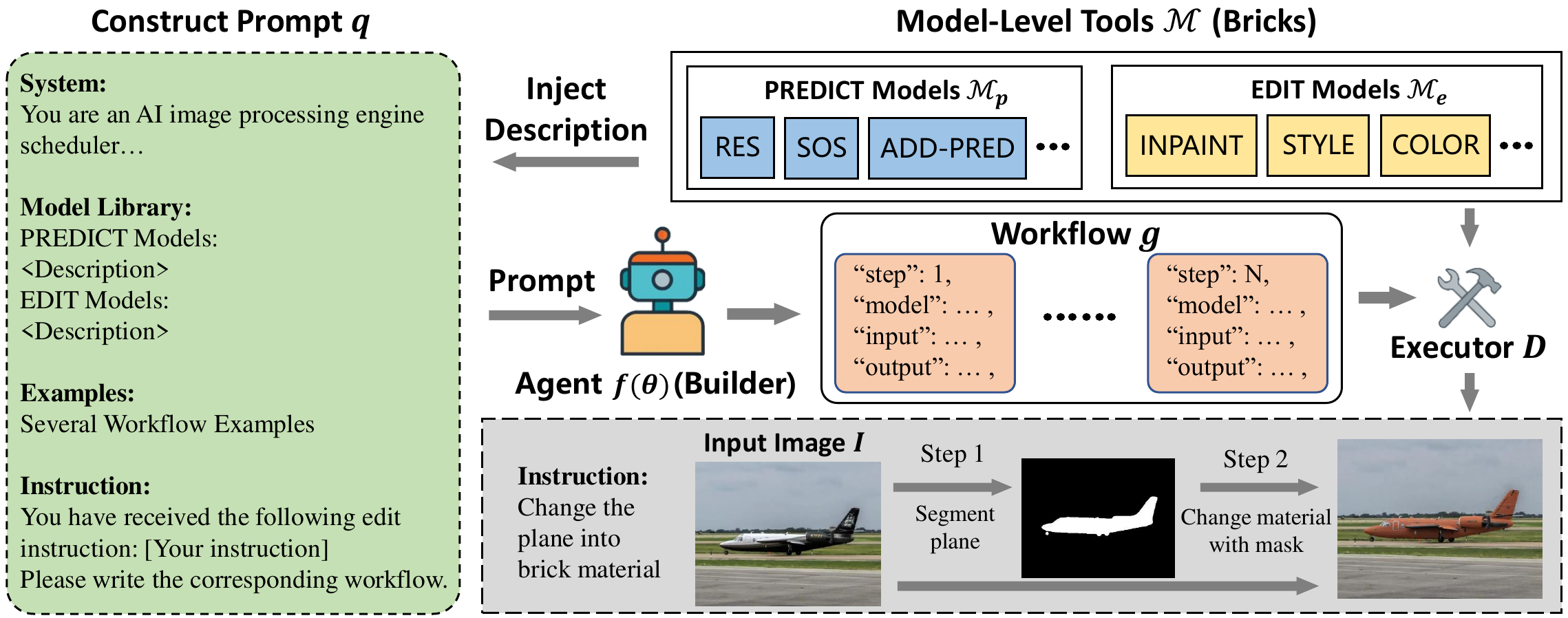}
\caption{Overall framework of Lego-Edit. Given an instruction and an input image, the Builder generates a tool invocation workflow. The Executor then executes this workflow, calling corresponding tools to generate the edited output image.}
\label{fig:overall_framework}
\end{figure*}

We introduce Lego-Edit, a framework designed for general instruction-based image editing. It uses Builder (an MLLM) to invoke Bricks (model-level tools) for flexibility and employs reinforcement learning (RL) to enhance the Builder's reasoning and tool composition ability. We first outline the overall framework, then detail the Builder's prompt structure. The following section describes tool classification, and the final section elaborates on our three-stage progressive RL training strategy for the Builder.

\subsection{Overall Framework}
\label{sec:overall_framework}

As illustrated in Fig.~\ref{fig:overall_framework}, our system comprises: 1) the Builder, an MLLM reasoning agent denoted as $f(\theta)$ that generates workflows, where $\theta$ denotes the model parameters; 2) an Executor $D$ that parses and executes workflows; and 3) the Bricks, an external model-level tool library $\mathcal{M} = {M_1, \dots, M_N}$ containing functions encapsulating models or logic processes, where $N$ is the total number of tools.

Given an input pair $(I, q)$ comprising a target image $I$ and an editing prompt $q$, the Builder $f(\theta)$, observing the state $s = (I, q)$, generates a reasoning trace, denoted as $Think$, and a JSON-formatted workflow $g$ based on its strategy $\pi_\theta(g \mid s)$. This workflow $g = (V, E)$ is a tool invocation graph. The vertex set $V = \{ M_1, ..., M_K \}$ represents the selected tool instances, with each $M_i \in \mathcal{M}$ and $1 \leq K \leq N$; here, $K$ is adaptively determined by the task complexity. The edge set $E \subseteq V \times V$ defines sequential dependencies, where an edge $(M_i, M_j)$ indicates that the input of $M_j$ depends on the output of $M_i$. The Executor $D$ then parses $g$, invokes the tools, and generates the edited image $I' = D(g, I)$.

\subsection{Prompt Structure}
\label{sec:prompt_structure}

Our curated Builder's input prompt format (left side of Fig.~\ref{fig:overall_framework}) has three key components:

\textbf{System Description and Invocation Constraints:} Defines capabilities, task scope, and valid tool parameter types via a system prompt.

\textbf{Available Tool List:} Each entry includes the model name, functional description, and invocation constraints detailing editing capabilities and requirements.

\textbf{Workflow Composition Examples:} Few-shot examples guiding valid workflow writing.

Following these, the prompt presents the editing instruction and directs the Builder to generate an editing workflow after reasoning.

\subsection{Model-Level Editing Tools}
\label{sec:model-level}

We constructed a fine-grained library of model-level image editing tools $\mathcal{M}$, where each tool represents a single model or function. Tools are categorized into two classes by whether they modify the image:

Predictive Models ($\mathcal{M}_p \subset \mathcal{M}$): Extract/process masks and regions to provide spatial constraints without altering pixels. Included tools are RES (segment specified objects), SOS (subject object segmentation), ADD-PRED (predict addition location), CAP-PRED (image captioning), INVERSE (invert the mask), and one additional tool.

Editing Models ($\mathcal{M}_e \subset \mathcal{M}$): Modify image content. Included tools are FILL (add object with given prompt or reference image), five specialized LoRA adapters trained on FILL: INPAINT (inpainting), POSE (human pose change), ENV (environment alteration), STYLE (style transfer), RCM (material/color change), and three additional tools.

To call a tool, the Builder needs to specify its name, input parameters, and output parameters. Complete details are in the supplementary material.

To prevent task confusion that can arise from joint training in end-to-end models (as in ICEdit \cite{zhang2025context}), we train independent LoRA adapters for each editing model. Furthermore, the Builder could precisely control edit scope using masks from $\mathcal{M}_p$, enabling more accurate editing.

\subsection{Three-Stage Progressive Reinforcement Learning Strategy}
\label{sec:3-stage}

To train the Builder $f(\theta)$ for effective tool composition, we employ a three-stage progressive RL strategy, gradually increasing task complexity and reducing reliance on ground truth data.

\subsubsection{Reinforcement Learning with GRPO}

We first introduce the Group Relative Policy Optimization (GRPO) \cite{shao2024deepseekmath} algorithm utilized in stages 2 and 3. For a given input $(I, q)$, the policy $\pi_{\theta}$ samples $G$ workflows $\{g_1, ..., g_G\}$. Each workflow $g_j$ receives a reward $r_j$ (defined per stage below). The relative advantage for each sample within the group is computed as:
\begin{equation}
A_j = \frac{r_j - \text{mean}(\{r_1, \ldots, r_G\})}{\text{std}(\{r_1, \ldots, r_G\})}, \quad j = 1, 2, \ldots, G
\end{equation}
The policy is updated by maximizing the GRPO objective:
\begin{equation}
\begin{aligned}
\mathcal{J}(\theta) &= \text{E}_{\begin{subarray}{l}
                                s \sim \mathcal{Q}, \\
                                \{g_j\}_{j=1}^G \sim \pi_{\theta_{\text{old}}}(g|s)
                               \end{subarray}} \left[ \frac{1}{G} \sum_{j=1}^G \min \left( r_j^{\text{ratio}} A_j, \text{clip}(r_j^{\text{ratio}}, 1 \pm \epsilon) A_j \right) - \beta \text{D}_{\text{KL}}(\pi_\theta \| \pi_{\text{ref}}) \right]
\end{aligned}
\end{equation}
where $\mathcal{Q}$ denotes the distribution of image-instruction pairs for sampling observation $s$, $\pi_{\theta_{\text{old}}}$ is the old policy before updating, $\pi_{\text{ref}}$ is a fixed reference policy, $r_j^{\text{ratio}} = \pi_\theta(g_j|s) / \pi_{\theta_{\text{old}}}(g_j|s)$, $\epsilon$ controls the clipping range, and $\beta$ weights the KL regularization $\text{D}_{\text{KL}}$ towards $\pi_{\text{ref}}$.

\subsubsection{Stage 1: Supervised Fine-Tuning (SFT)}

We adapt the Builder to image editing domain using SFT on data from several specific tasks. Each sample $(I, q)$ is paired with expert-generated reasoning traces $Think_{\text{GT}}$ and ground truth workflows $g_{\text{GT}}$. The learning target is denoted as the concatenated sequence $l = [Think_{\text{GT}}, g_{\text{GT}}]$. The model is trained to minimize the negative log-likelihood:
\begin{equation}
L_{\text{SFT}} = -\sum_{t=1}^{T} \log p_\theta(l_t \mid I, q, l_{<t})
\end{equation}
where $T$ is the total length of $l$, and $p_\theta$ is the model's conditional next-token distribution.

\begin{figure}
\centering
\includegraphics[width=0.65\linewidth]{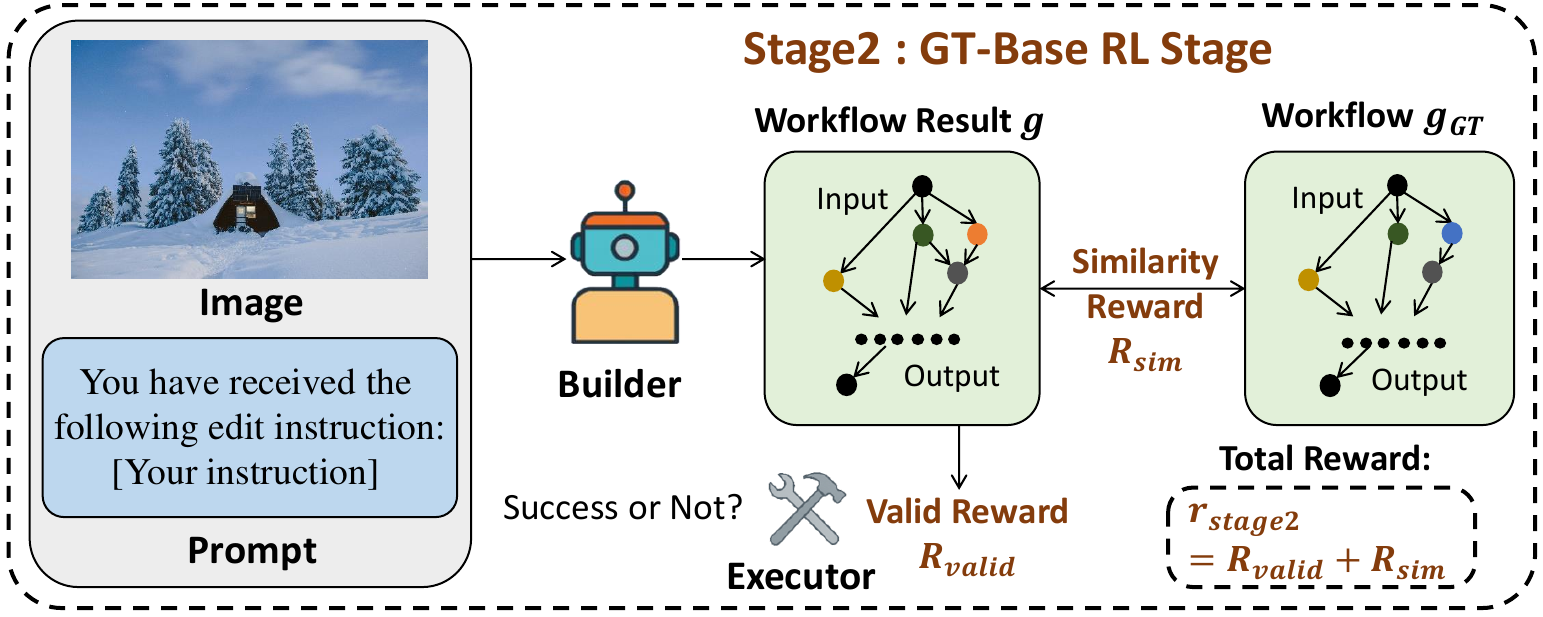}
\caption{Illustration of the reward design adopted in Stage 2 (GT-based RL training).}
\label{fig:rl}
\end{figure}

\subsubsection{Stage 2: Refinement with GT-Based Rewards}

Building on SFT, Stage 2 continues training on specific tasks using only $(I, q)$ pairs and $g_{\text{GT}}$. The workflow generated by Builder is denoted $g$. We adopt two rewards at this stage:

Valid Reward ($R_{\text{valid}}$): It penalizes non-executable workflows as follows:
\begin{equation}
R_{\text{valid}} = \begin{cases}
0, & \text{if $D$ successfully executes $g$} \\
-1, & \text{otherwise}
\end{cases}
\label{eq:valid}
\end{equation}

Similarity Reward ($R_{\text{sim}}$): It measures alignment between $g$ and the expert workflow $g_{\text{GT}}$ using hierarchical graph matching. The depth of nodes in $g$ is calculated based on inverse topological order, which means the depth of the output node is 0. Matched nodes $M$ are found per depth layer using the Hungarian algorithm based on node similarity with a threshold of 0.6. The node similarity $\text{sim}_{\text{node}}$ is calculated by averaging the indicator of whether the same model is used and the proportion of identical parameters. $R_{\text{sim}}$ combines node coverage and average matched node similarity:
\begin{equation}
\begin{aligned}
R_{\text{sim}} &= 0.5 \cdot \frac{|M|}{\max(|V_g|, |V_{g_{\text{GT}}}|)} 
+ 0.5 \cdot \frac{1}{|M|} \sum_{(i,j) \in M} \text{sim}_{\text{node}}(i,j)
\end{aligned}
\end{equation}
where $V_g$ and $V_{g_{\text{GT}}}$ denote the node sets of the generated workflow $g$ and the ground truth workflow $g_{\text{GT}}$  respectively.

The total reward is: $r_{\text{stage2}} = R_{\text{valid}} + R_{\text{sim}}$.

\begin{figure}
\centering
\includegraphics[width=0.65\linewidth]{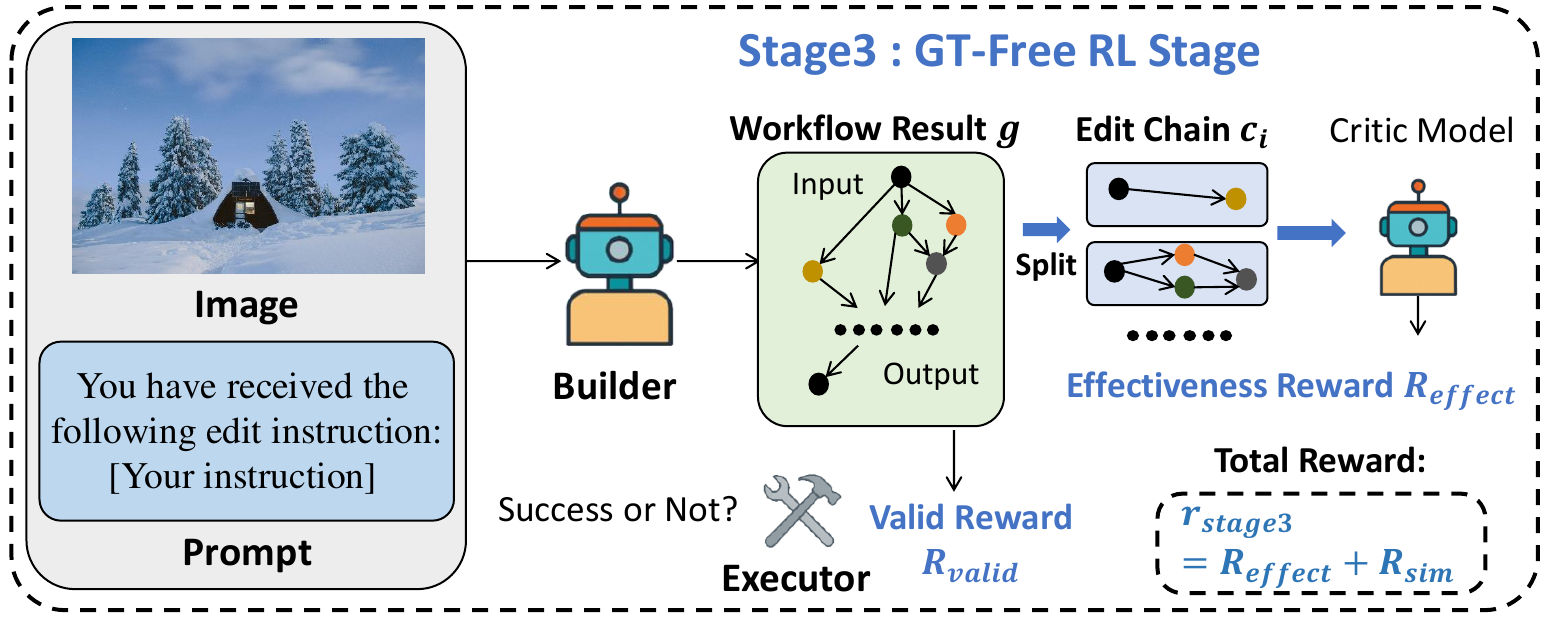}
\caption{Illustration of the reward design adopted in Stage 3 (GT-free RL training).}
\label{fig:rl}
\end{figure}

\subsubsection{Stage 3: Generalization with GT-Free Critic Rewards}

Stage 3 targets generalization to open-domain instructions using only $(I, q)$ pairs. We employ the same valid reward $R_{\text{valid}}$ as Eq.~\ref{eq:valid} and another effectiveness reward to provide feedback without a ground truth workflow:

Effectiveness Reward ($R_{\text{effect}}$): It uses an MLLM critic model to assess semantic alignment between the workflow's effect and instruction $q$. Workflows are decomposed into several editing chains $c_i$, each of which contains only one editing model in $\mathcal{M}_e$ to perform actual editing. The MLLM critic would abstract each chain's effect into a meta-edit description $m_i$ and then evaluate the description set $\{m_i\}$ against $q$. Specifically, the LLM must determine whether to remove existing editing chains or add new editing chains to better achieve the instruction. It must specify the number of chains to remove ($N_{\text{remove}}$), the number of chains to add ($N_{\text{add}}$), and the content of new chains. $R_{\text{effect}}$ applies a penalty defined as follows:
\begin{equation}
R_{\text{effect}} = 1-0.5\cdot(N_{\text{add}} + N_{\text{remove}})
\end{equation}
The total reward is: $r_{\text{stage3}} = R_{\text{valid}} + R_{\text{effect}}$.

\section{Experiments}
We first demonstrate the model's zero-shot capability, followed by extensive experiments validating the superiority of our framework on image editing benchmarks. Then we analyze the performance improvements and other advantages introduced by model-level tools, and finally demonstrate the Builder's improved generalization and performance enabled by reinforcement learning.

\subsection{Implementation Details}
\label{sec:4.1}
\noindent\textbf{Builder:} Our Builder is based on MiMo-VL-7B \cite{coreteam2025mimovltechnicalreport} and undergoes full-parameter fine-tuning in bf16 using a progressive three-stage curriculum: (1) 500 image-text pairs (instructions, thoughts and workflows), (2) 20K pairs (instructions and workflows), and (3) 50K pairs (instructions), all sourced from OmniEdit \cite{wei2024omniedit}. The generation of thoughts and workflows, along with the critic model used in stage 3 training, are both based on Qwen2.5-VL-72B \cite{bai2025qwen2}. Each stage trains for 1 epoch with AdamW (lr = $1\mathrm{e}{-5}$, $\beta = (0.9, 0.999)$, weight decay = 0.01), without warmup or decay (batch size 8, image size $448 \times 448$). Data construction details are provided in the supplementary material.

\noindent\textbf{Predictive Tools:} RES utilizes EVF-SAM \cite{zhang2024evf}, trained from scratch on 200K MS COCO \cite{lin2014microsoft} samples (lr = $1\mathrm{e}{-4}$, batch size 64, resolution $512 \times 512$, 10K iterations) with BCE and Dice loss.
SOS employs U2Net \cite{qin2020u2}, trained on 5K DIS \cite{qin2022highly} with identical loss and optimization settings (batch size 24, 100 epochs, BCE loss only). Both models initialize without pretrained weights.
ADD-PRED and CAP-PRED share a Qwen2-VL-2B backbone \cite{wang2024qwen2}. CAP-PRED directly utilizes the base model's inherent captioning capability. ADD-PRED is fine-tuned on 50K OmniEdit samples for addition/removal region prediction, formulated as bounding box regression between the source and target images; training uses 1 epoch and a learning rate of $1\mathrm{e}{-5}$.

\noindent\textbf{Editing Tools:} We adopt ICEdit's \cite{zhang2025context} framework (FLUX-1 Fill \cite{fluxfill2024} backbone with LoRA fine-tuning at rank=32) but implement five specialized adapters for individual tasks, rather than the multi-task adapter. Each adapter is trained on 10K task-specific samples curated from OmniEdit \cite{wei2024omniedit} and MagicBrush \cite{zhang2023magicbrush} via VIEScore \cite{ku2023viescore} assessment, using a global batch size of 8 for 10K steps at 768×768 resolution.

All the experiments utilized 8$\times$NVIDIA H20 GPUs for training. 
We adopt DeepSpeed ZeRO-3 \cite{rajbhandari2020zero} to accelerate training.

\subsection{Evaluation Settings}
\label{sec:4.2}
To ensure authoritative evaluation, we benchmark our method on two widely adopted datasets: GEdit-Bench (606 samples) \cite{liu2025step1x} and ImgEdit (811 samples) \cite{ye2025imgedit}, 
known for complex editing instructions and high-quality imagery. 
Following standard protocols, we employ VIEScore executed by GPT-4o \cite{hurst2024gpt} as our primary metric. To ensure fairness and reproducibility, we fix the random seed to 0 and perform single-shot inference for all evaluations.
On GEdit-Bench using a single H20 GPU, our Builder takes 3.5 seconds and the slowest Tool 2.7 seconds, resulting in a total pipeline latency of $\sim$ 7.2 seconds, versus $>$ 25 seconds for the end-to-end method Bagel under identical settings.

\begin{figure*}[!h]
\centering
\includegraphics[width=\linewidth]{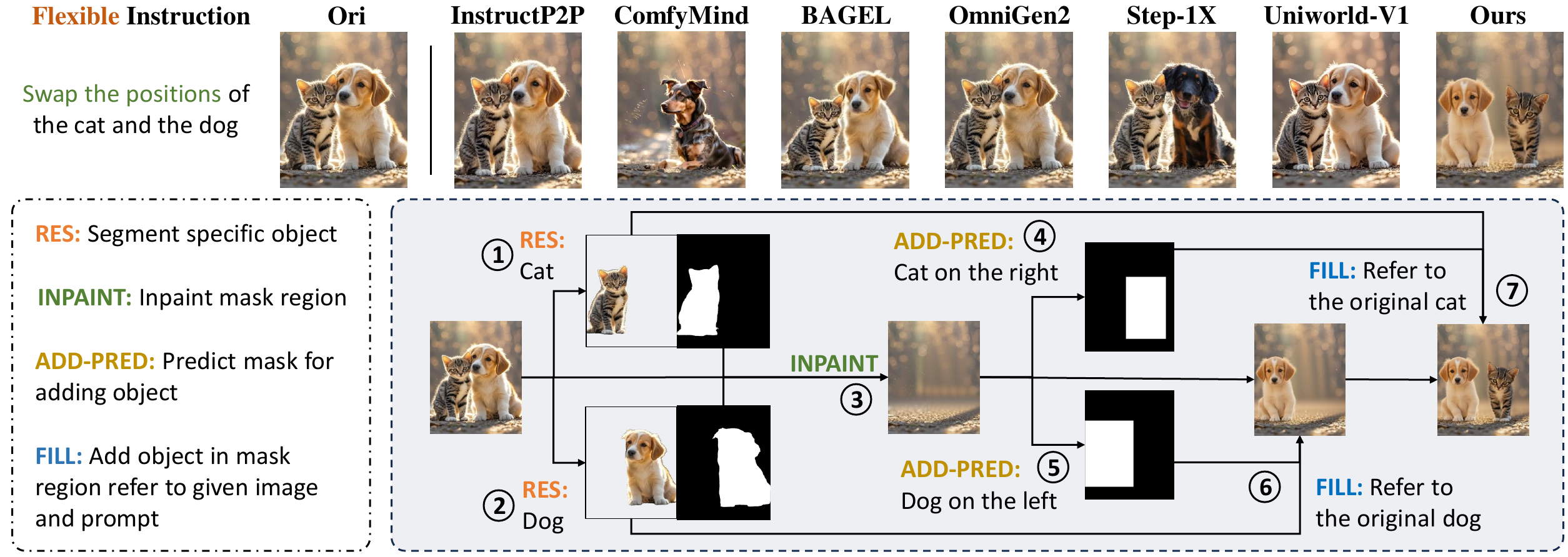}
\caption{Comparison with other methods on complex edits (top) and our Tool composition workflow (bottom).}
\label{fig:img6}
\end{figure*}

\subsection{Zero-Shot Capability of the Builder}
\noindent\textbf{Zero-Shot Complex Edits with Flexible Tool Composition:} Fig.~\ref{fig:img6} presents a visual comparison of editing results on flexible instructions, alongside the Builder's tool composition process. For the ``swap" instruction, although the Builder was not explicitly trained on this task, it effectively decomposes the instruction into atomic operations by first removing object A using RES and INPAINT, then inserting object B via ADD-PRED and FILL. This example shows its ability to compose specialized tools for flexible editing instructions, which enables complex edits beyond the reach of end-to-end or curated-pipeline models.

\begin{figure*}[!h]
\centering
\includegraphics[width=\linewidth]{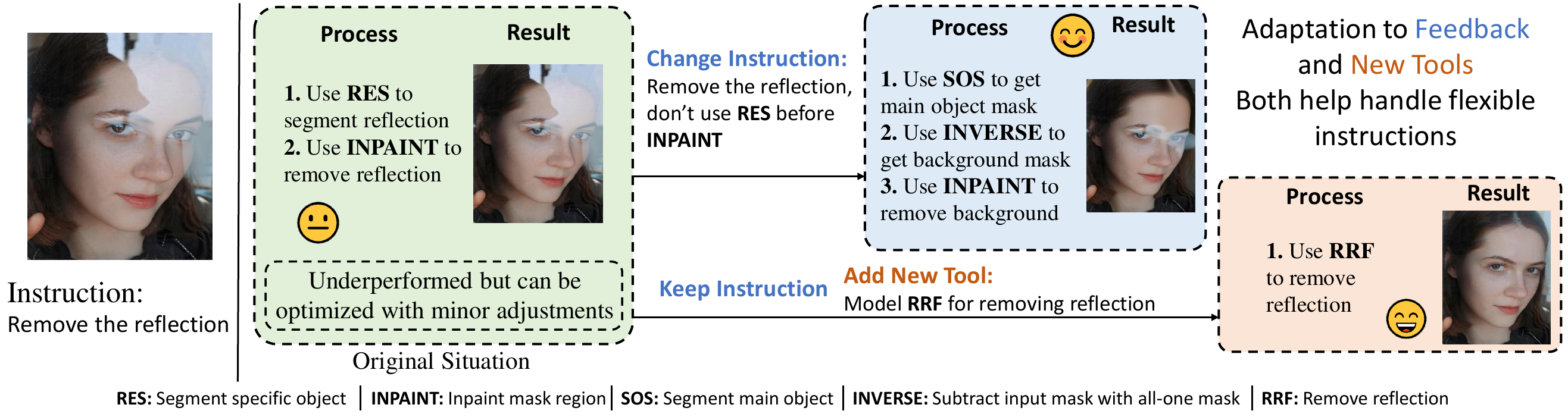}
\caption{Example of Zero-Shot adaptation via feedback and Tool Insertion in reflection removal.}
\label{fig:img5}
\end{figure*}

\noindent \textbf{Zero-Shot Adaptation to Feedback and New Tools:} 
Fig.~\ref{fig:img5} demonstrates the Builder's adaptability to user feedback and new tools without retraining. For the reflection removal task, the Builder's initial workflow (RES and INPAINT) failed because RES could not segment reflections effectively. Users can provide direct instructions, such as ``don't use RES before INPAINT" to prevent it. Guided by this feedback, the Builder revises its workflow: it uses SOS for foreground segmentation, INVERSE to infer the background, and then INPAINT to remove part of the reflection. Additionally, users can introduce a dedicated reflection-removal tool (RRF), which the Builder readily adopts to solve the task effectively. 
This illustrates the system's adaptability to extend capabilities by integrating new tools or incorporating user feedback, all without modifying the Builder.

\begin{figure}[h!]
    \centering
    \begin{minipage}[b]{0.36\linewidth}
        \centering
        \resizebox{\linewidth}{!}{%
        \begin{tabular}{l|ccc}
        \toprule
        \textbf{Model} & \textbf{G\_SC}$\uparrow$ & \textbf{G\_PQ}$\uparrow$ & \textbf{G\_O}$\uparrow$ \\
        \midrule
        Instruct-P2P
        & 3.58 & 5.49 & 3.68 \\
        MagicBrush
        & 4.68 & 5.66 & 4.52 \\
        AnyEdit
        & 3.18 & 5.82 & 3.21 \\
        OmniGen
        & 5.96 & 5.89 & 5.06 \\
        Step1X-Edit
        & \underline{7.13} & 7.00 & 6.44 \\
        BAGEL
        & \textbf{7.36} & 6.83 & \underline{6.52} \\
        UniWorld-V1
        & 4.93 & \underline{7.43} & 4.85 \\
        ComfyMind
        & 2.67 & 5.93 & 2.61 \\
        OmniGen2
        & - & - & 6.42 \\
        \midrule
        \textbf{Ours}
        & 5.99 & \textbf{7.45} & \textbf{6.64} \\
        \bottomrule
        \end{tabular}%
        }
        \caption{Quantitative evaluation on GEdit-Bench-EN. All metrics are reported as higher-is-better ($\uparrow$).}
        \label{tab:gedit}
    \end{minipage}
    \hfill
    \begin{minipage}[b]{0.62\linewidth}
        \centering
        \vspace{-\baselineskip} 
        \includegraphics[width=0.75\linewidth, height=0.35\textheight, keepaspectratio]{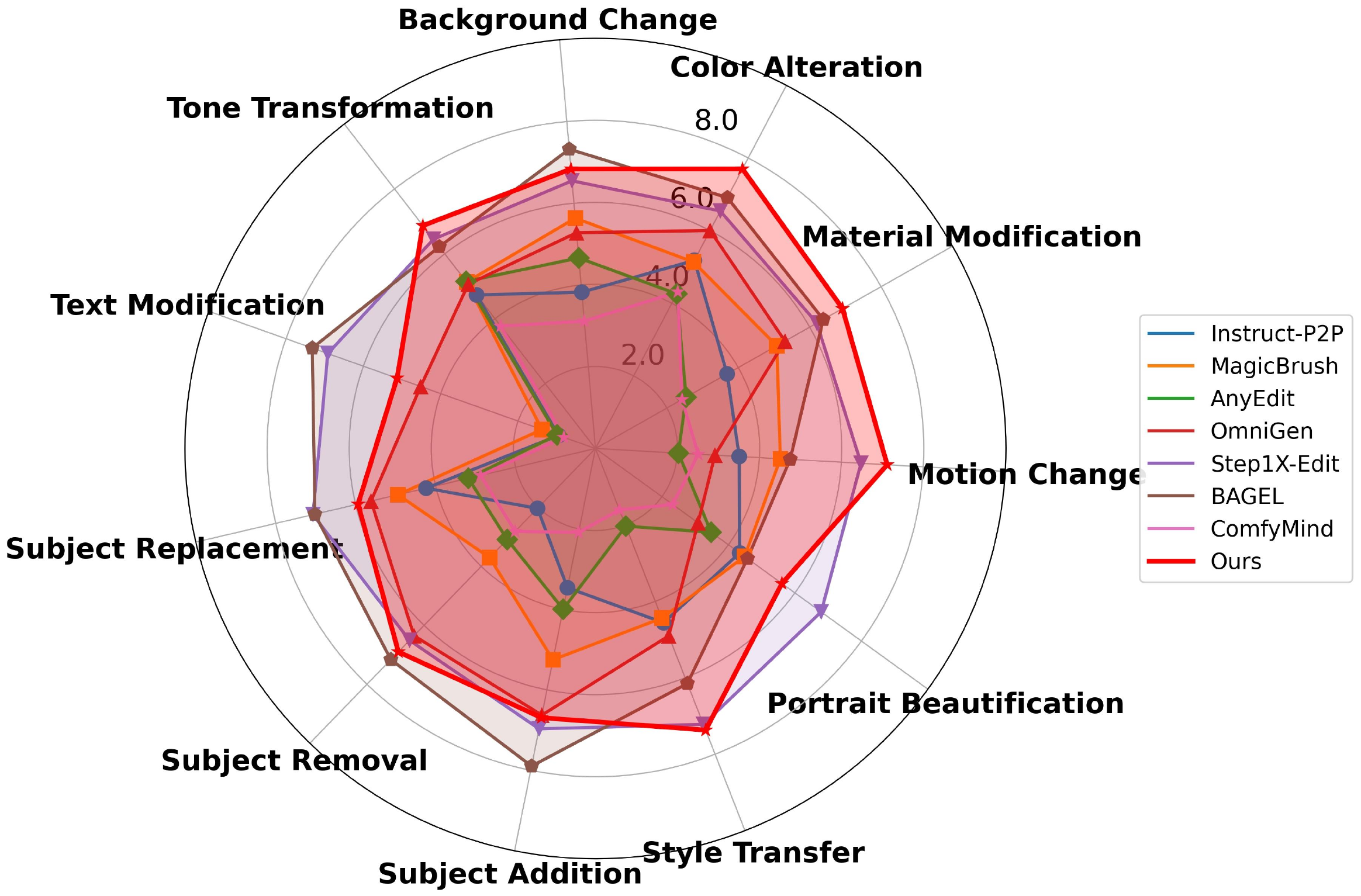}
        \caption{VIEScore of Each Sub-task in GEdit-Bench-En. All the results are evaluated by GPT-4o.}
        \label{fig:lidar}
    \end{minipage}
\end{figure}

\subsection{Comparison with State-of-the-Art}
\label{sec:4.3}
\textbf{GEdit-Bench:}  As shown in Tab.~\ref{tab:gedit}, our method achieves the highest image preservation score on GEdit-Bench-EN, with G\_PQ = 7.7, and delivers the best overall performance with G\_O = 6.64, outperforming all competing approaches. 
As shown in Fig.~\ref{fig:lidar}, Lego-Edit excels in fine-grained sub-tasks such as color change and material replacement. This precision is attributed to the Builder composing RES to execute these tasks with minimal impact on non-target regions, as detailed in the section on Effectiveness of Tool Composition.
Compared to traditional manual-workflow API agents like ComfyMind, which depend on pre-scripted pipelines, Lego-Edit achieves notable performance gains via a capable Builder and flexible orchestration of specialized Tools.

\begin{table*}[!h]
  \centering
  \resizebox{\linewidth}{!}
  {
    \begin{tabular}{l|ccccccccc|c}
    \toprule
    \textbf{Model} & \textbf{Add} & \textbf{Adjust} & \textbf{Extract} & \textbf{Replace} & \textbf{Remove} & \textbf{Style} & \textbf{Action} & \textbf{Hybrid} & \textbf{Background} & \textbf{Overall$\uparrow$}\\
    \midrule
    MagicBrush \cite{zhang2023magicbrush} & 2.72 & 1.47 & 1.31 & 1.89 & 1.57 & 2.49 & 1.39 & 1.80 & 2.03  & 1.85\\ 
    Instruct-P2P \cite{brooks2023instructpix2pix} & 2.29 & 1.79 & 1.33 & 1.93 & 1.49 & 3.54 & 1.51 & 1.48 & 1.67 & 1.89\\
    AnyEdit \cite{yu2025anyedit} & 3.12 & 2.66 & 1.82 & 2.71 & 2.34 & 3.27 & 3.31 & 2.07 & 2.37 & 2.63\\
    UltraEdit \cite{zhao2024ultraedit} & 3.63 & 3.01 & 2.02 & 3.13 & 1.71 & 3.69 & 3.57 & 2.33 & \underline{3.31} & 2.93 \\
    Step1X-Edit \cite{liu2025step1x} & \textbf{3.90} & 3.13 & 1.87 & 3.45 & 2.61 & 4.44 & 3.43 & 2.52 & 3.19 & 3.17\\
    BAGEL \cite{deng2025emerging} & 3.55 & 3.30 & 1.56 & 3.38 & 2.44 & 4.24 & \underline{4.29} & 2.55 & 3.22 & 3.17 \\
    UniWorld-V1 \cite{lin2025uniworld} & \underline{3.86} & \underline{3.70} & \underline{2.23} & \underline{3.49} & \textbf{3.54} & 4.22 & 3.44 & \underline{3.13} & 2.76 & 3.37\\
    ComfyMind \cite{guo2025comfymind} & 1.45 & 3.14 & 2.21 & 3.43 & 2.81 & 2.66 & 2.74 & 0.57 & 2.26 & 2.63\\
    OmniGen2 \cite{wu2025omnigen2} & 3.57 & 3.06 & 1.77 & \textbf{3.74} & 3.20 &  \textbf{4.81} & \textbf{4.68} & 2.52 & \textbf{3.57} & \underline{3.44}\\
    \midrule
    \textbf{Ours} & 3.67 & \textbf{3.82} & \textbf{2.47} & 3.22 &  \underline{3.39} &  \underline{4.47} & 4.01 & \textbf{3.18} & 3.24  & \textbf{3.50}\\
    \bottomrule
    \end{tabular}
  }
  \caption{Quantitative evaluation on ImgEdit-Bench. All metrics are reported as higher-is-better ($\uparrow$).}
  \label{tab:imgbench}
\end{table*}

\noindent  \textbf{ImgEdit-Bench:} Lego-Edit maintains top performance on ImgBench, 
achieving the highest overall score (3.50) among all compared methods. The detailed results are shown in Tab.~\ref{tab:imgbench}.
Crucially, our framework dominates the most challenging Hybrid Editing sub-task (3.18). This success validates our proposition that the Builder can parse composite instructions into atomic sub-tasks and dynamically generate workflows to coordinate specialized Tools.

\begin{figure*}[!h]
\centering
\includegraphics[width=\linewidth]{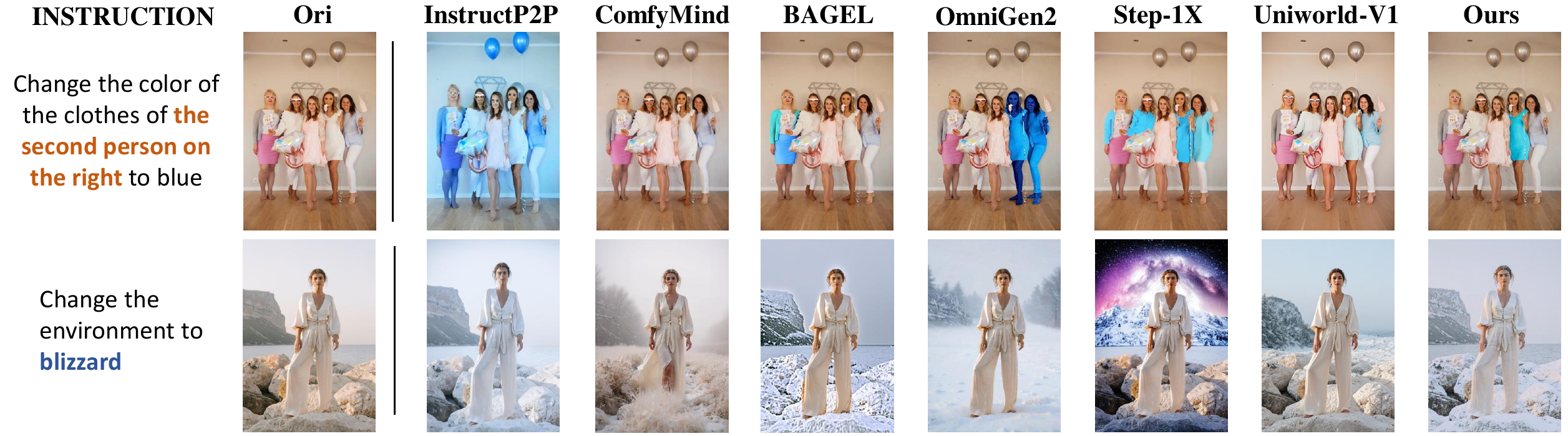}
\caption{Qualitative results compared to other methods.}
\label{fig:img8}
\end{figure*}

\noindent  \textbf{Qualitative Results:} As illustrated in Fig.~\ref{fig:img8}, our method outperforms other approaches in both edit accuracy and visual realism. Edits are well-aligned with the intended regions and maintain high perceptual quality. More comparative results are provided in the supplementary material. 

\subsection{Ablation Study on Tools}
\label{sec:4.4}
\textbf{Necessity of Task-Specialized Tools.} 
To validate Lego-Edit's design, we compare its task-specialized architecture with a unified alternative using identical settings. Three separate LoRA adapters are trained on 10K samples per task, while the unified model uses a combined 30K dataset. As shown in Tab.~\ref{tab:icedit}, specialized models outperform the unified one (e.g., 6.83 vs. 5.94 in color alteration). Increasing LoRA rank in the unified model brings no gain. Qualitative results (provided in supplementary material) reveal frequent task confusion in the unified setup, highlighting the importance of specialization for editing fidelity.

\begin{table}[h]
  \centering
  \resizebox{0.64\linewidth}{!}
  {
    \begin{tabular}{l|cccc}
    \toprule
    \multirow{2}{*}{\textbf{Model}} & \multirow{2}{*}{\textbf{Rank}} & \textbf{Color} & \textbf{Style} & \textbf{Tone} \\
    &  & \textbf{Alteration} & \textbf{Transfer} & \textbf{Transformation} \\
    \midrule
    Multi-task LoRA  & 32 & 5.94 & 5.47 & 5.56 \\
    Multi-task LoRA & 64 & 5.42 & 5.43 & 5.57 \\
    Single-task LoRA & 32 & \textbf{6.83} & \textbf{6.75}  & \textbf{6.63}\\
    \bottomrule
    \end{tabular}
  }
  \caption{Comparison between Multi-task LoRA and Single-task LoRA on GEdit-Bench-EN, we report the {G\_O} metric.}
  \label{tab:icedit}
\end{table}

\subsection{Ablation Studies on Builder}
\label{sec:4.5}

\begin{table}[htbp]
  \centering
  \resizebox{0.68\linewidth}{!}
  {
    \begin{tabular}{l|cc|ccc}
    \toprule
    \multirow{2}{*}{\textbf{Agent Model}} & \textbf{Simple} & \textbf{Complex} &  \multirow{2}{*}{\textbf{G\_SC}$\uparrow$} & \multirow{2}{*}{\textbf{G\_PQ}$\uparrow$} & \multirow{2}{*}{\textbf{G\_O}$\uparrow$} \\
     & \textbf{\%Pass} & \textbf{\%Pass} & & &  \\
    \midrule
    GPT-4o & 75.6 & 53.8 & 4.25 & 6.64 & 5.12\\
    MiMo-VL-7B & 57.2 & 47.7 & 3.24 & 4.87 & 3.88 \\
    Builder-SFT & 80.6 & 55.9 & 4.83 & 6.94 & 5.89  \\
    Builder-RL w/ GT & \textbf{100} & 83.6 & 5.58 & 7.32 & 6.17 \\
    Builder-RL w/o GT & \textbf{100} & \textbf{99.0} & \textbf{5.99} & \textbf{7.45} & \textbf{6.64}  \\
    \bottomrule
    \end{tabular}
  }
  \caption{ Performance is evaluated by execution success rate, indicating error-free syntax, and VIEScore, assessing editing quality. Success rate comprises Simple \%Pass on original instructions and Complex \%Pass on three GPT-4o-generated variants per instruction. }
  \label{tab:grpo}
\end{table}

\textbf{Effectiveness of Reinforcement Learning Training.} 
Ablations on GEdit-Bench in Tab.~\ref{tab:grpo} show the effectiveness of our progressive RL training. Starting from Builder-SFT, which outperforms basemodel MiMo-VL-7B and powerful MLLM GPT-4o on simple and complex success rates and VIEScores, subsequent RL training with ground truth (Builder-RL w/ GT) achieves 100\% simple and 83.6\% complex success with better VIEScores. Final GT-free RL training (Builder-RL w/o GT) maintains 100\% simple success, boosts complex success to 99.0\%, and achieves the highest VIEScores. This complex task improvement, despite identical simple accuracy, indicates more accurate parameter output and enhanced compositional understanding of tool interfaces, enabling reliable execution under flexible instructions.

\noindent \textbf{Effectiveness of Tool Composition.} To evaluate the impact of explicit Tool composition in Lego-Edit, we examine its performance across GEdit-Bench sub-tasks. 
As shown in Tab.~\ref{tab:resplus}, integrating RES segmentation masks with RCM increases G\_O for color alteration and material modification, highlighting the benefits of RES's spatial control. 
Similarly, incorporating CAP-PRED text descriptions into STYLE boosts G\_O in style transfer, as textual cues enforce semantic alignment between stylized outputs and the source image. 
These results demonstrate that Tool composition improves performance, confirming that Lego-Edit coordinates specialized tools to achieve superior editing performance.

\begin{table}[htbp]
  \centering
  \resizebox{0.52\linewidth}{!}
  {
    \begin{tabular}{l|c|c}
    \toprule
    \textbf{Task} & \textbf{Compose Pattern} & \textbf{G\_O} \\
    \midrule
    \multirow{2}{*}{\textbf{Color Alteration}} & RCM & 6.83 \\
     & RCM + RES & \textbf{7.70} \\
    \midrule
    \multirow{2}{*}{\textbf{Material Modification}} & RCM & 6.08 \\
     & RCM + RES & \textbf{6.92} \\
    \midrule
    \multirow{2}{*}{\textbf{Style Transfer}} & STYLE & 6.75\\
     & STYLE + CAP-PRED & \textbf{7.37} \\
    \bottomrule
    \end{tabular}
  }
  \caption{Comparison between separately adopted tools and with Builder compose on GEdit-Bench-EN.}
  \label{tab:resplus}
\end{table}

\section{Conclusion}

In this paper, we introduce Lego-Edit, a novel agent-based framework for generative instruction-based image editing. It employs an RL-finetuned MLLM (Builder) to orchestrate model-level editing tools (Bricks), enabled by two key innovations: fine-grained tools allowing flexible composition and precise control, and a three-stage progressive RL training strategy enhancing reasoning and tool organization abilities via GT-free feedback on open-domain instructions. Extensive experiments demonstrate Lego-Edit's state-of-the-art performance on GEdit-Bench and ImgBench, showcasing superior accuracy and generalization in handling flexible requests and integrating new tools without retraining. Future work will expand the tool set for broader capabilities and explore agent-based feedback for better robustness.

\bibliography{iclr2026_conference}

\begin{thebibliography}{37}
\providecommand{\natexlab}[1]{#1}
\providecommand{\url}[1]{\texttt{#1}}
\expandafter\ifx\csname urlstyle\endcsname\relax
  \providecommand{\doi}[1]{doi: #1}\else
  \providecommand{\doi}{doi: \begingroup \urlstyle{rm}\Url}\fi

\bibitem[Bai et~al.(2025)Bai, Chen, Liu, Wang, Ge, Song, Dang, Wang, Wang, Tang, et~al.]{bai2025qwen2}
Shuai Bai, Keqin Chen, Xuejing Liu, Jialin Wang, Wenbin Ge, Sibo Song, Kai Dang, Peng Wang, Shijie Wang, Jun Tang, et~al.
\newblock Qwen2. 5-vl technical report.
\newblock \emph{arXiv preprint arXiv:2502.13923}, 2025.

\bibitem[Brooks et~al.(2023)Brooks, Holynski, and Efros]{brooks2023instructpix2pix}
Tim Brooks, Aleksander Holynski, and Alexei~A Efros.
\newblock Instructpix2pix: Learning to follow image editing instructions.
\newblock In \emph{Proceedings of the IEEE/CVF conference on computer vision and pattern recognition}, pp.\  18392--18402, 2023.

\bibitem[Cai et~al.(2025)Cai, Chen, Chen, Li, Long, Pan, Qiu, Zhang, Gao, Xu, et~al.]{cai2025hidream}
Qi~Cai, Jingwen Chen, Yang Chen, Yehao Li, Fuchen Long, Yingwei Pan, Zhaofan Qiu, Yiheng Zhang, Fengbin Gao, Peihan Xu, et~al.
\newblock Hidream-i1: A high-efficient image generative foundation model with sparse diffusion transformer.
\newblock \emph{arXiv preprint arXiv:2505.22705}, 2025.

\bibitem[Deng et~al.(2025)Deng, Zhu, Li, Gou, Li, Wang, Zhong, Yu, Nie, Song, et~al.]{deng2025emerging}
Chaorui Deng, Deyao Zhu, Kunchang Li, Chenhui Gou, Feng Li, Zeyu Wang, Shu Zhong, Weihao Yu, Xiaonan Nie, Ziang Song, et~al.
\newblock Emerging properties in unified multimodal pretraining.
\newblock \emph{arXiv preprint arXiv:2505.14683}, 2025.

\bibitem[Du et~al.(2024)Du, Wei, and Zhang]{du2024anytool}
Yu~Du, Fangyun Wei, and Hongyang Zhang.
\newblock Anytool: Self-reflective, hierarchical agents for large-scale api calls.
\newblock \emph{arXiv preprint arXiv:2402.04253}, 2024.

\bibitem[Gao et~al.(2025)Gao, Lu, Zhou, Chu, Zhang, Jia, Zhang, Fan, and Zhang]{gao2025eraseanything}
Daiheng Gao, Shilin Lu, Wenbo Zhou, Jiaming Chu, Jie Zhang, Mengxi Jia, Bang Zhang, Zhaoxin Fan, and Weiming Zhang.
\newblock Eraseanything: Enabling concept erasure in rectified flow transformers.
\newblock In \emph{Forty-second International Conference on Machine Learning}, 2025.

\bibitem[Guo et~al.(2025)Guo, Xu, Wang, Lin, Zhou, Zhang, Su, and Chen]{guo2025comfymind}
Litao Guo, Xinli Xu, Luozhou Wang, Jiantao Lin, Jinsong Zhou, Zixin Zhang, Bolan Su, and Ying-Cong Chen.
\newblock Comfymind: Toward general-purpose generation via tree-based planning and reactive feedback.
\newblock \emph{arXiv preprint arXiv:2505.17908}, 2025.

\bibitem[Han et~al.(2024)Han, Jiang, Pan, Zhang, Mao, Xie, Liu, and Zhou]{han2024ace}
Zhen Han, Zeyinzi Jiang, Yulin Pan, Jingfeng Zhang, Chaojie Mao, Chenwei Xie, Yu~Liu, and Jingren Zhou.
\newblock Ace: All-round creator and editor following instructions via diffusion transformer.
\newblock \emph{arXiv preprint arXiv:2410.00086}, 2024.

\bibitem[Hertz et~al.(2022)Hertz, Mokady, Tenenbaum, Aberman, Pritch, and Cohen-Or]{hertz2022prompt}
Amir Hertz, Ron Mokady, Jay Tenenbaum, Kfir Aberman, Yael Pritch, and Daniel Cohen-Or.
\newblock Prompt-to-prompt image editing with cross attention control.
\newblock \emph{arXiv preprint arXiv:2208.01626}, 2022.

\bibitem[Huang et~al.(2025{\natexlab{a}})Huang, Wang, Yang, Lu, Yuan, Han, Hou, Zhang, Hong, Zhao, et~al.]{huang2025illume+}
Runhui Huang, Chunwei Wang, Junwei Yang, Guansong Lu, Yunlong Yuan, Jianhua Han, Lu~Hou, Wei Zhang, Lanqing Hong, Hengshuang Zhao, et~al.
\newblock Illume+: Illuminating unified mllm with dual visual tokenization and diffusion refinement.
\newblock \emph{arXiv preprint arXiv:2504.01934}, 2025{\natexlab{a}}.

\bibitem[Huang et~al.(2024)Huang, Xie, Wang, Yuan, Cun, Ge, Zhou, Dong, Huang, Zhang, et~al.]{huang2024smartedit}
Yuzhou Huang, Liangbin Xie, Xintao Wang, Ziyang Yuan, Xiaodong Cun, Yixiao Ge, Jiantao Zhou, Chao Dong, Rui Huang, Ruimao Zhang, et~al.
\newblock Smartedit: Exploring complex instruction-based image editing with multimodal large language models.
\newblock In \emph{Proceedings of the IEEE/CVF Conference on Computer Vision and Pattern Recognition}, pp.\  8362--8371, 2024.

\bibitem[Huang et~al.(2025{\natexlab{b}})Huang, Ji, Rajan, Cai, Xiao, Hu, and Lee]{huang2025visualtoolagent}
Zeyi Huang, Yuyang Ji, Anirudh~Sundara Rajan, Zefan Cai, Wen Xiao, Junjie Hu, and Yong~Jae Lee.
\newblock Visualtoolagent (vista): A reinforcement learning framework for visual tool selection.
\newblock \emph{arXiv preprint arXiv:2505.20289}, 2025{\natexlab{b}}.

\bibitem[Hurst et~al.(2024)Hurst, Lerer, Goucher, Perelman, Ramesh, Clark, Ostrow, Welihinda, Hayes, Radford, et~al.]{hurst2024gpt}
Aaron Hurst, Adam Lerer, Adam~P Goucher, Adam Perelman, Aditya Ramesh, Aidan Clark, AJ~Ostrow, Akila Welihinda, Alan Hayes, Alec Radford, et~al.
\newblock Gpt-4o system card.
\newblock \emph{arXiv preprint arXiv:2410.21276}, 2024.

\bibitem[Ku et~al.(2023)Ku, Jiang, Wei, Yue, and Chen]{ku2023viescore}
Max Ku, Dongfu Jiang, Cong Wei, Xiang Yue, and Wenhu Chen.
\newblock Viescore: Towards explainable metrics for conditional image synthesis evaluation.
\newblock \emph{arXiv preprint arXiv:2312.14867}, 2023.

\bibitem[Labs(2024{\natexlab{a}})]{flux2024}
Black~Forest Labs.
\newblock Flux.
\newblock \url{https://github.com/black-forest-labs/flux}, 2024{\natexlab{a}}.

\bibitem[Labs(2024{\natexlab{b}})]{fluxfill2024}
Black~Forest Labs.
\newblock Flux.1-fill-dev.
\newblock \url{https://huggingface.co/black-forest-labs/FLUX.1-Fill-dev}, 2024{\natexlab{b}}.

\bibitem[Lin et~al.(2025)Lin, Li, Cheng, Niu, Ye, He, Yuan, Yu, Wang, Ge, et~al.]{lin2025uniworld}
Bin Lin, Zongjian Li, Xinhua Cheng, Yuwei Niu, Yang Ye, Xianyi He, Shenghai Yuan, Wangbo Yu, Shaodong Wang, Yunyang Ge, et~al.
\newblock Uniworld: High-resolution semantic encoders for unified visual understanding and generation.
\newblock \emph{arXiv preprint arXiv:2506.03147}, 2025.

\bibitem[Lin et~al.(2014)Lin, Maire, Belongie, Hays, Perona, Ramanan, Doll{\'a}r, and Zitnick]{lin2014microsoft}
Tsung-Yi Lin, Michael Maire, Serge Belongie, James Hays, Pietro Perona, Deva Ramanan, Piotr Doll{\'a}r, and C~Lawrence Zitnick.
\newblock Microsoft coco: Common objects in context.
\newblock In \emph{European conference on computer vision}, pp.\  740--755. Springer, 2014.

\bibitem[Liu et~al.(2025)Liu, Han, Xing, Yin, Wang, Cheng, Liao, Wang, Fu, Han, et~al.]{liu2025step1x}
Shiyu Liu, Yucheng Han, Peng Xing, Fukun Yin, Rui Wang, Wei Cheng, Jiaqi Liao, Yingming Wang, Honghao Fu, Chunrui Han, et~al.
\newblock Step1x-edit: A practical framework for general image editing.
\newblock \emph{arXiv preprint arXiv:2504.17761}, 2025.

\bibitem[Qin et~al.(2020)Qin, Zhang, Huang, Dehghan, Zaiane, and Jagersand]{qin2020u2}
Xuebin Qin, Zichen Zhang, Chenyang Huang, Masood Dehghan, Osmar~R Zaiane, and Martin Jagersand.
\newblock U2-net: Going deeper with nested u-structure for salient object detection.
\newblock \emph{Pattern recognition}, 106:\penalty0 107404, 2020.

\bibitem[Qin et~al.(2022)Qin, Dai, Hu, Fan, Shao, and Van~Gool]{qin2022highly}
Xuebin Qin, Hang Dai, Xiaobin Hu, Deng-Ping Fan, Ling Shao, and Luc Van~Gool.
\newblock Highly accurate dichotomous image segmentation.
\newblock In \emph{European Conference on Computer Vision}, pp.\  38--56. Springer, 2022.

\bibitem[Qin et~al.(2023)Qin, Liang, Ye, Zhu, Yan, Lu, Lin, Cong, Tang, Qian, et~al.]{qin2023toolllm}
Yujia Qin, Shihao Liang, Yining Ye, Kunlun Zhu, Lan Yan, Yaxi Lu, Yankai Lin, Xin Cong, Xiangru Tang, Bill Qian, et~al.
\newblock Toolllm: Facilitating large language models to master 16000+ real-world apis.
\newblock \emph{arXiv preprint arXiv:2307.16789}, 2023.

\bibitem[Rajbhandari et~al.(2020)Rajbhandari, Rasley, Ruwase, and He]{rajbhandari2020zero}
Samyam Rajbhandari, Jeff Rasley, Olatunji Ruwase, and Yuxiong He.
\newblock Zero: Memory optimizations toward training trillion parameter models.
\newblock In \emph{SC20: International Conference for High Performance Computing, Networking, Storage and Analysis}, pp.\  1--16. IEEE, 2020.

\bibitem[Shao et~al.(2024)Shao, Wang, Zhu, Xu, Song, Bi, Zhang, Zhang, Li, Wu, et~al.]{shao2024deepseekmath}
Zhihong Shao, Peiyi Wang, Qihao Zhu, Runxin Xu, Junxiao Song, Xiao Bi, Haowei Zhang, Mingchuan Zhang, YK~Li, Yang Wu, et~al.
\newblock Deepseekmath: Pushing the limits of mathematical reasoning in open language models.
\newblock \emph{arXiv preprint arXiv:2402.03300}, 2024.

\bibitem[Team et~al.(2025)Team, Yue, Lin, Song, Wang, Ren, Gu, Li, Li, Zhao, Li, et~al.]{coreteam2025mimovltechnicalreport}
Core Team, Zihao Yue, Zhenru Lin, Yifan Song, Weikun Wang, Shuhuai Ren, Shuhao Gu, Shicheng Li, Peidian Li, Liang Zhao, Lei Li, et~al.
\newblock Mimo-vl technical report, 2025.
\newblock URL \url{https://arxiv.org/abs/2506.03569}.

\bibitem[Wang et~al.(2024)Wang, Bai, Tan, Wang, Fan, Bai, Chen, Liu, Wang, Ge, et~al.]{wang2024qwen2}
Peng Wang, Shuai Bai, Sinan Tan, Shijie Wang, Zhihao Fan, Jinze Bai, Keqin Chen, Xuejing Liu, Jialin Wang, Wenbin Ge, et~al.
\newblock Qwen2-vl: Enhancing vision-language model's perception of the world at any resolution.
\newblock \emph{arXiv preprint arXiv:2409.12191}, 2024.

\bibitem[Wei et~al.(2024)Wei, Xiong, Ren, Du, Zhang, and Chen]{wei2024omniedit}
Cong Wei, Zheyang Xiong, Weiming Ren, Xeron Du, Ge~Zhang, and Wenhu Chen.
\newblock Omniedit: Building image editing generalist models through specialist supervision.
\newblock In \emph{The Thirteenth International Conference on Learning Representations}, 2024.

\bibitem[Wu et~al.(2025)Wu, Zheng, Yan, Xiao, Luo, Wang, Li, Jiang, Liu, Zhou, et~al.]{wu2025omnigen2}
Chenyuan Wu, Pengfei Zheng, Ruiran Yan, Shitao Xiao, Xin Luo, Yueze Wang, Wanli Li, Xiyan Jiang, Yexin Liu, Junjie Zhou, et~al.
\newblock Omnigen2: Exploration to advanced multimodal generation.
\newblock \emph{arXiv preprint arXiv:2506.18871}, 2025.

\bibitem[Xue et~al.(2025)Xue, Lu, Huang, Wang, Ouyang, and Bai]{xue2025comfybench}
Xiangyuan Xue, Zeyu Lu, Di~Huang, Zidong Wang, Wanli Ouyang, and Lei Bai.
\newblock Comfybench: Benchmarking llm-based agents in comfyui for autonomously designing collaborative ai systems.
\newblock In \emph{Proceedings of the Computer Vision and Pattern Recognition Conference}, pp.\  24614--24624, 2025.

\bibitem[Ye et~al.(2025)Ye, He, Li, Lin, Yuan, Yan, Hou, and Yuan]{ye2025imgedit}
Yang Ye, Xianyi He, Zongjian Li, Bin Lin, Shenghai Yuan, Zhiyuan Yan, Bohan Hou, and Li~Yuan.
\newblock Imgedit: A unified image editing dataset and benchmark.
\newblock \emph{arXiv preprint arXiv:2505.20275}, 2025.

\bibitem[Yu et~al.(2025)Yu, Chow, Yue, Pan, Wu, Wan, Li, Tang, Zhang, and Zhuang]{yu2025anyedit}
Qifan Yu, Wei Chow, Zhongqi Yue, Kaihang Pan, Yang Wu, Xiaoyang Wan, Juncheng Li, Siliang Tang, Hanwang Zhang, and Yueting Zhuang.
\newblock Anyedit: Mastering unified high-quality image editing for any idea.
\newblock In \emph{Proceedings of the Computer Vision and Pattern Recognition Conference}, pp.\  26125--26135, 2025.

\bibitem[Zhang et~al.(2023)Zhang, Mo, Chen, Sun, and Su]{zhang2023magicbrush}
Kai Zhang, Lingbo Mo, Wenhu Chen, Huan Sun, and Yu~Su.
\newblock Magicbrush: A manually annotated dataset for instruction-guided image editing.
\newblock \emph{Advances in Neural Information Processing Systems}, 36:\penalty0 31428--31449, 2023.

\bibitem[Zhang et~al.(2024)Zhang, Cheng, Zhu, Hu, Liu, Liu, Ran, Chen, Liu, and Wang]{zhang2024evf}
Yuxuan Zhang, Tianheng Cheng, Lianghui Zhu, Rui Hu, Lei Liu, Heng Liu, Longjin Ran, Xiaoxin Chen, Wenyu Liu, and Xinggang Wang.
\newblock Evf-sam: Early vision-language fusion for text-prompted segment anything model.
\newblock \emph{arXiv preprint arXiv:2406.20076}, 2024.

\bibitem[Zhang et~al.(2025)Zhang, Xie, Lu, Yang, and Yang]{zhang2025context}
Zechuan Zhang, Ji~Xie, Yu~Lu, Zongxin Yang, and Yi~Yang.
\newblock In-context edit: Enabling instructional image editing with in-context generation in large scale diffusion transformer.
\newblock \emph{arXiv preprint arXiv:2504.20690}, 2025.

\bibitem[Zhao et~al.(2024)Zhao, Ma, Chen, Si, Wu, An, Yu, Zhang, Li, and Chang]{zhao2024ultraedit}
Haozhe Zhao, Xiaojian~Shawn Ma, Liang Chen, Shuzheng Si, Rujie Wu, Kaikai An, Peiyu Yu, Minjia Zhang, Qing Li, and Baobao Chang.
\newblock Ultraedit: Instruction-based fine-grained image editing at scale.
\newblock \emph{Advances in Neural Information Processing Systems}, 37:\penalty0 3058--3093, 2024.

\bibitem[Zheng et~al.(2024{\natexlab{a}})Zheng, Gou, Kil, Sun, and Su]{zheng2024gpt}
Boyuan Zheng, Boyu Gou, Jihyung Kil, Huan Sun, and Yu~Su.
\newblock Gpt-4v (ision) is a generalist web agent, if grounded.
\newblock \emph{arXiv preprint arXiv:2401.01614}, 2024{\natexlab{a}}.

\bibitem[Zheng et~al.(2024{\natexlab{b}})Zheng, Li, Liu, Liu, Luan, and Wang]{zheng2024toolrerank}
Yuanhang Zheng, Peng Li, Wei Liu, Yang Liu, Jian Luan, and Bin Wang.
\newblock Toolrerank: Adaptive and hierarchy-aware reranking for tool retrieval.
\newblock \emph{arXiv preprint arXiv:2403.06551}, 2024{\natexlab{b}}.

\end{thebibliography}
\bibliographystyle{iclr2026_conference}

\appendix

\newpage

\section{Details of Model-Level Tools}

We provide detailed specifications of all model-level tools in Tab.~\ref{tab:tools}, including their names, functions, inputs, and outputs.

\begin{table*}[h!]
  \centering
  \resizebox{\linewidth}{!}
  {
    \begin{tabular}{c|m{2.5cm}|p{9.0cm}|p{3.0cm}|m{2.0cm}}
    \toprule
    \textbf{Tool Type} & \textbf{Tool Name} & \textbf{Function} & \textbf{Input} & \textbf{Output} \\
    \midrule
     & \multirow{2}{*}{RES} & \multirow{2}{*}{Segment object specified by prompt.} & Image[image],  & Mask[mask],  \\
    &&&Str[prompt]&Image[image] \\
    \cline{2-5}
    & \multirow{2}{*}{SOS} & \multirow{2}{*}{Segment salient objects in image.} & \multirow{2}{*}{Image[image]} & Mask[mask], \\
    &&&& Image[image]\\
    \cline{2-5}
    & \multirow{3}{*}{ADD-PRED} & \multirow{3}{9.0cm}{Predict optimal position to add target object (specified by prompt). If mask provided, position must be within mask.} & Image[image],  & \multirow{3}{*}{Mask[mask]} \\
    &&&Str[prompt],&\\
    &&& Mask[mask]&\\
    \cline{2-5}
    \textbf{Predictive}& \multirow{5}{*}{CAP-PRED} & \multirow{5}{9.0cm}{Generate English caption describing the image for FLUX model conditioning. For image expansion tasks, requires expansion ratios; outputs caption, expanded image, and expansion mask. Otherwise, ratios null, outputs caption only (image/mask null).} & Image[image], &   \\
    &&&Float[left\_ratio],&Str[caption],\\
    \textbf{Models}&&&Float[right\_ratio],&Image[image],\\
    &&&Float[top\_ratio],&Mask[mask]\\
    &&&Float[bottom\_ratio]&\\
    \cline{2-5}
    & \multirow{4}{*}{INVERSE} & \multirow{4}{9.0cm}{Subtract mask2 from mask1 or image2 from image1. If mask1 is null, subtract mask2 from a full mask.} & Mask[mask1],  &   \\
    &&&Mask[mask2],&Mask[mask],\\
    &&&Image[image1],& Image[image] \\
    &&&Image[image2]&\\
    \cline{2-5}
    & BBOX & Compute bounding box mask from input mask. & Mask[mask] & Mask[mask] \\
    \midrule
     & \multirow{4}{*}{FILL} & \multirow{4}{9.0cm}{Generate content within mask region based on prompt. Not for object color/material replacement.} & Image[image], &  \\
    &&&Mask[mask],&\multirow{2}{*}{Image[image]}\\
    &&&Str[prompt],&\\
    &&&Image[preimage]&\\
    \cline{2-5}
    & \multirow{2}{*}{FASTINPAINT} & \multirow{2}{9.0cm}{Perform fast inpainting and output an effectiveness score.} & Image[image],  & Image[image],  \\
    &&&Mask[mask]&Float[score]\\
    \cline{2-5}
    & \multirow{4}{*}{INPAINT} & \multirow{4}{9.0cm}{Fill background in mask region, using preimage as reference and requiring a quality score threshold.} & Image[image], & \multirow{4}{*}{Image[image]} \\
    &&&Mask[mask],&\\
    &&&Image[preimage],&\\
    &&&Float[score]&\\
    \cline{2-5}
    & \multirow{3}{*}{RCM} & \multirow{3}{9.0cm}{Replace color or material of object within mask.} & Image[image],   & \multirow{3}{*}{Image[image]} \\
    &&&Mask[mask],&\\
    \textbf{Editing}&&&Str[prompt]&\\
    \cline{2-5}
    & \multirow{4}{*}{STYLE} & \multirow{4}{9.0cm}{Convert style of entire image or object within mask to specified style (e.g., 'anime style').} & Image[image], & \multirow{4}{*}{Image[image]} \\
    &&&Mask[mask],&\\
    \textbf{Models}&&&Str[prompt],&\\
    &&&Str[style]&\\
    \cline{2-5}
    & \multirow{2}{*}{ENV} & \multirow{2}{9.0cm}{Alter environment (e.g., weather, time of day) of an object.} & Image[image],  & \multirow{2}{*}{Image[image]} \\
    &&&Str[prompt]&\\
    \cline{2-5}
    & \multirow{2}{*}{POSE} & \multirow{2}{9.0cm}{Modify posture or expression of an object.} & Image[image],  & \multirow{2}{*}{Image[image]} \\
    &&&Str[prompt]&\\
    \cline{2-5}
    & \multirow{4}{*}{COMPOSE} & \multirow{4}{9.0cm}{Combine two masks or images. Overlapping regions use the second input.} & Mask[mask1], & \\
    &&&Mask[mask2],&Mask[mask],\\
    &&&Image[image1],&Image[image]\\
    &&&Image[image2]&\\
    \cline{2-5}
    & \multirow{3}{*}{RESIZE} & \multirow{3}{9.0cm}{Resize the valid region of input mask or image by a given scaling ratio.} & Mask[mask], & \multirow{3}{2.0cm}{Mask[mask], Image[image]}\\
    &&&Image[image],&\\
    &&&Float[ratio]&\\
    \bottomrule
    \end{tabular}
  }
  \caption{Specification of Model-Level Tools: Names, functions, inputs, and outputs.}
  \label{tab:tools}
\end{table*}

\section{Additional Qualitative Comparisons with State-of-the-Art Methods}

\begin{figure*}[htbp]
\centering
\includegraphics[width=\linewidth]{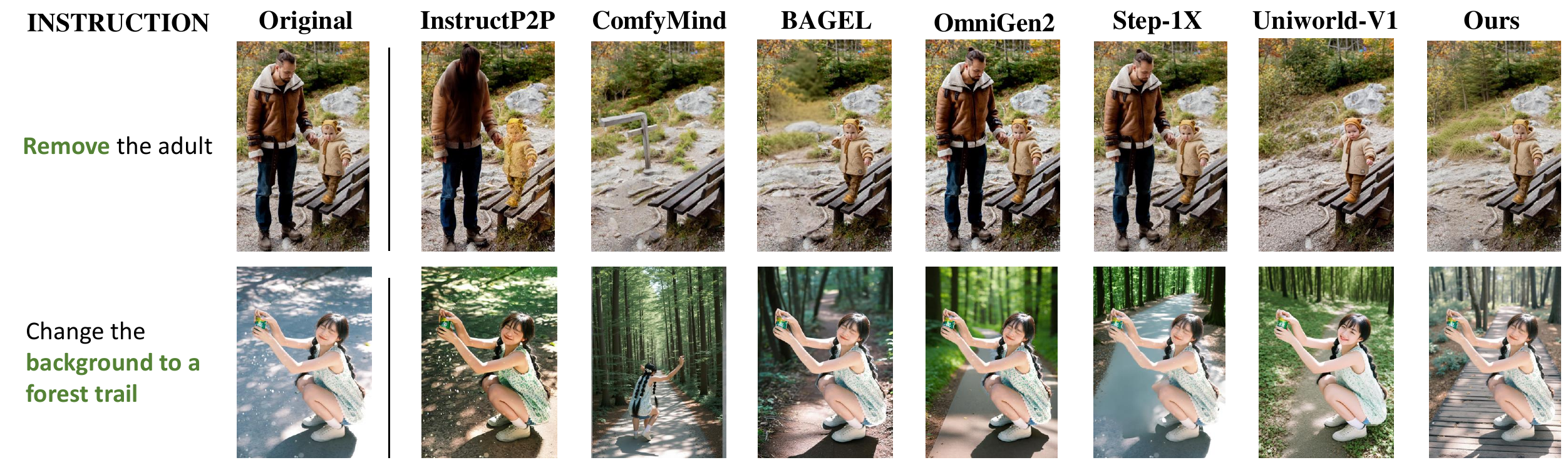}
\caption{Qualitative comparison with state-of-the-art methods.}
\label{fig:img1}
\end{figure*}

We present extended qualitative comparisons with state-of-the-art (SOTA) methods in Fig.~\ref{fig:img1}. The comparison methods include InstructP2P \cite{brooks2023instructpix2pix}, ComfyMind \cite{guo2025comfymind}, BAGEL \cite{deng2025emerging}, OmniGen2 \cite{wu2025omnigen2}, Step-1X \cite{liu2025step1x}, Uniworld-V1 \cite{lin2025uniworld}.

\begin{figure}[!h]
\centering
\includegraphics[width=0.8\linewidth]{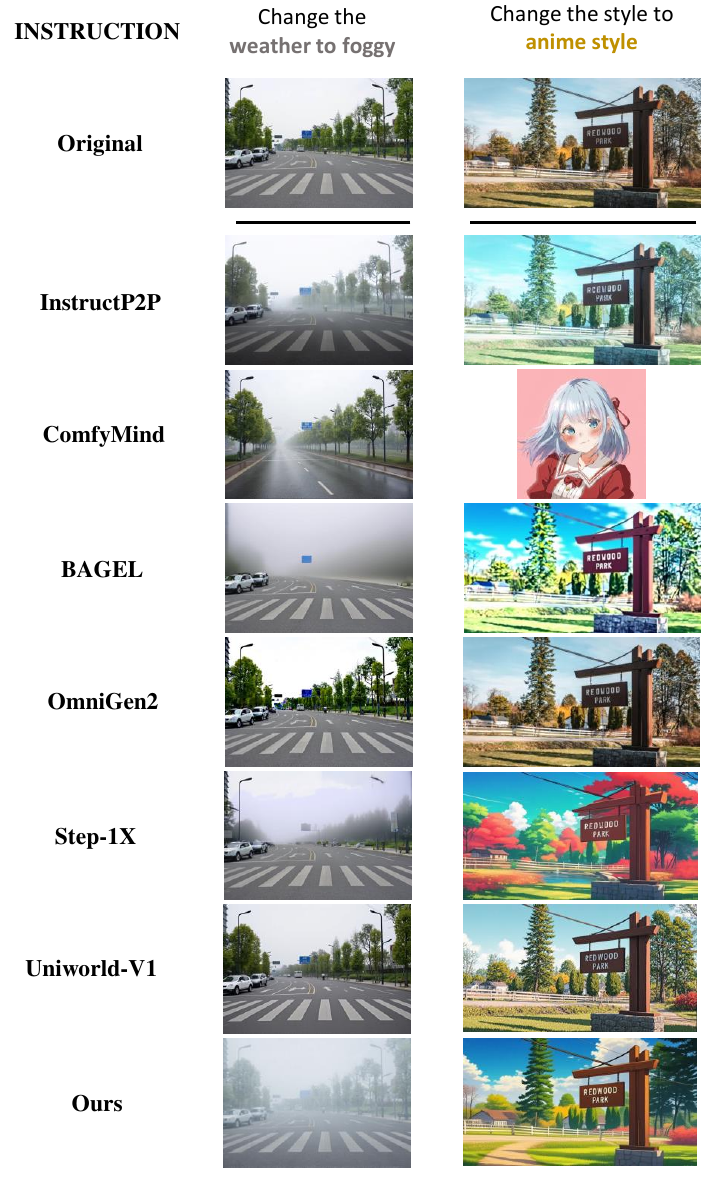}
\caption{Qualitative results for image editing at varying aspect ratios.}
\label{fig:img2}
\end{figure}

\noindent Further examples demonstrating performance across different aspect ratios are shown in Fig.~\ref{fig:img2}.

\begin{figure*}[htbp]
\centering
\includegraphics[width=\linewidth]{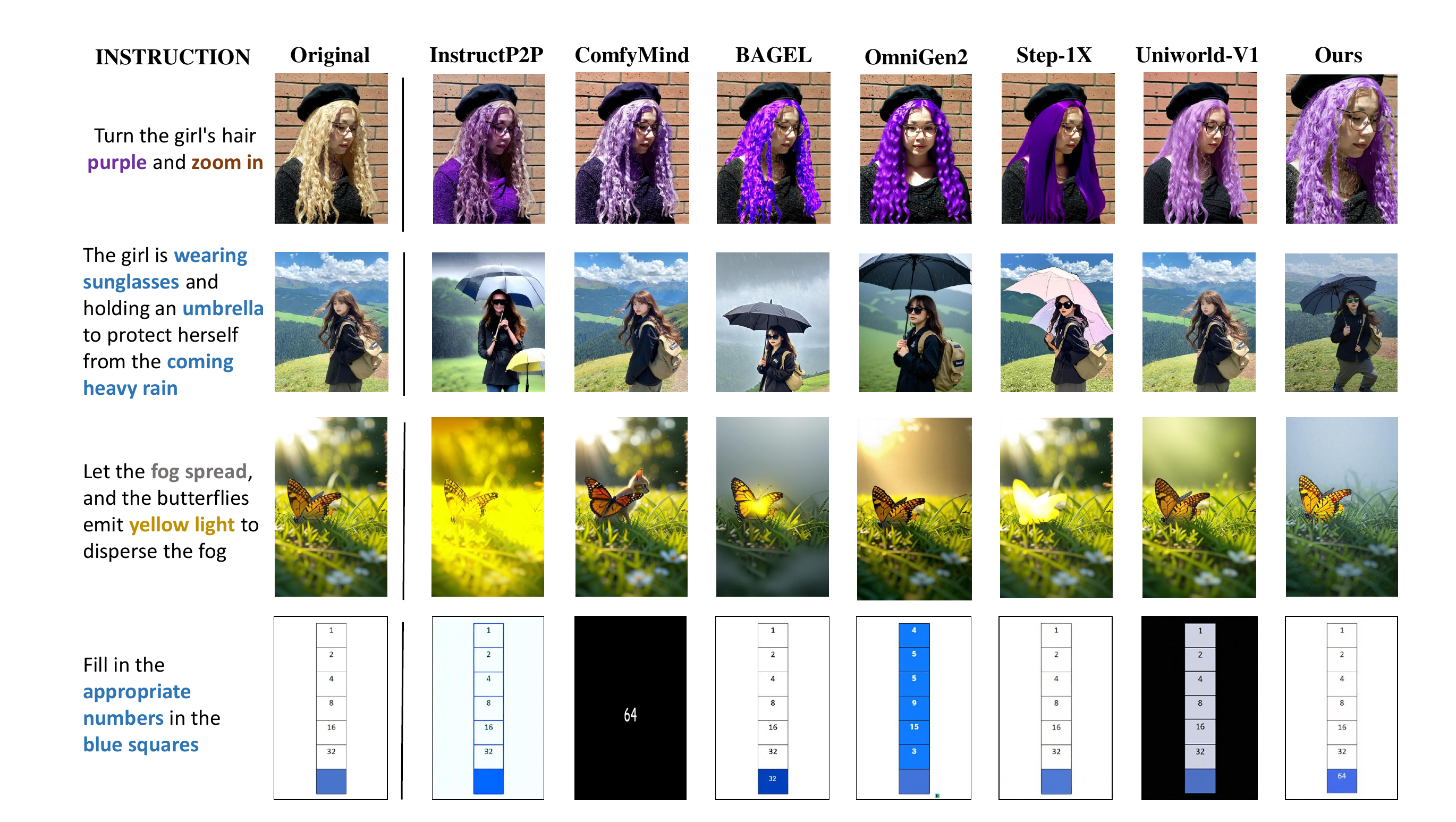}
\caption{Qualitative results of handling complex and flexible editing instructions.}
\label{fig:img3}
\end{figure*}

\noindent Fig.~\ref{fig:img3} showcases results achieved with more complex and flexible instructions.

\section{Qualitative Analysis of Task Confusion in Joint Training}

\begin{figure}[h!]
\centering
\includegraphics[width=0.6\linewidth]{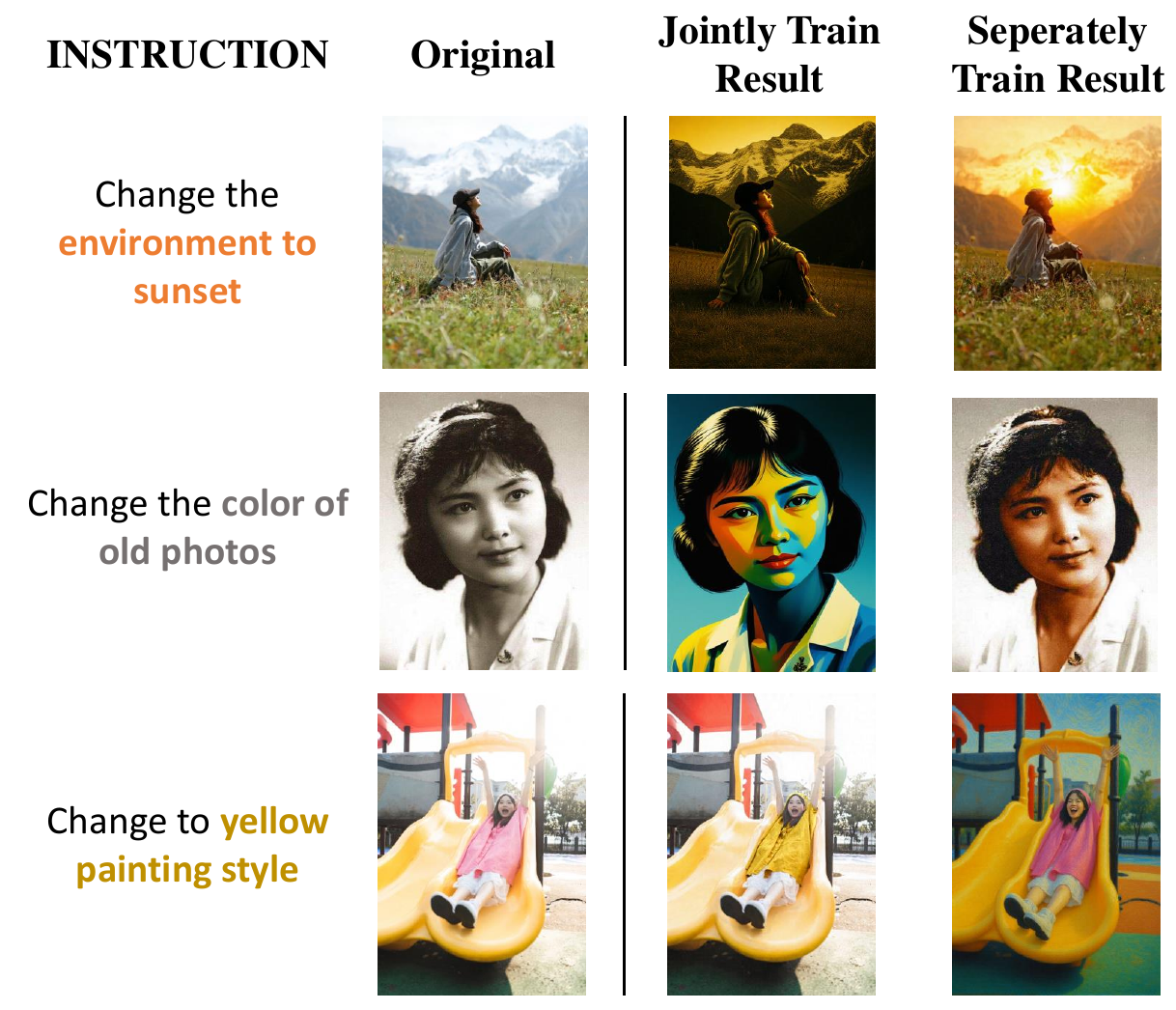}
\caption{Qualitative examples of task confusion arising from joint model training.}
\label{fig:img4}
\end{figure}

As discussed in the main paper, joint training approaches such as ICEdit \cite{zhang2025context} can suffer from task confusion. While quantitative evidence is presented therein, Fig.~\ref{fig:img4} provides qualitative examples illustrating this phenomenon.

\begin{figure}[h!]
\centering
\includegraphics[width=0.6\linewidth]{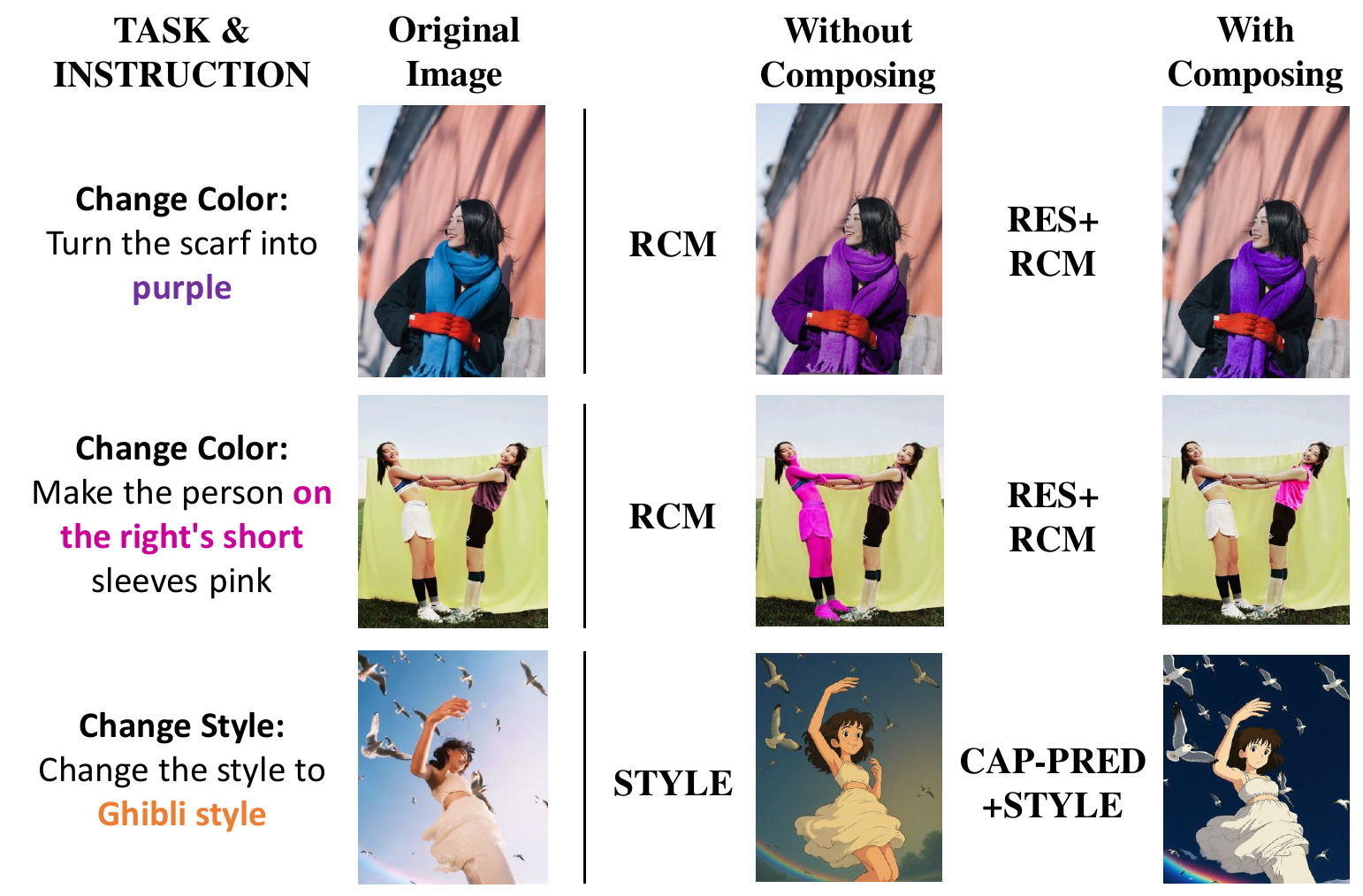}
\caption{Qualitative improvements achieved through strategic composition of model-level tools.}
\label{fig:img5}
\end{figure}

\section{Qualitative Benefits of Tool Composition}

The main paper presents quantitative results demonstrating the performance gains achieved by composing specific tools (e.g., RES for color/style change, CAP-PRED for style change). Fig.~\ref{fig:img5} offers corresponding qualitative evidence of these improvements.

\section{Complete Prompt Template for the Builder Agent}

The full prompt used to configure the Builder agent is provided in Listing~\ref{lst:main-prompt}.

\begin{lstlisting}[style=promptstyle,caption={Complete Prompt for the Builder Agent},label={lst:main-prompt}]
**System Role**
You are an AI image processing engine scheduler, responsible for converting the natural language instructions provided by the user into executable multi-model collaboration process json. All inputs must come from the initial parameters or the output of the previous model or your own language decomposition and translation.

**Processing rules**
1. Input traceability principle
- Each model parameter can only be:
a. Initial image, it can be an image list composed of many images
b. Text in user instructions
c. Output of previous steps
d. The result of your own language analysis and translation of the text in user instructions

2. Process generation steps
a. Extract the operation object and action from the user instruction and analyze it in combination with the image content
b. After analysis, select the corresponding model for each operation, ensure that the model you choose and the parameters you input meet the requirements of the model
c. Establish a cross-model data dependency chain, make sure the output of each step is used in subsequent processes, otherwise this step is redundant.

**Input and output types**
There are only four types of input and output
1. Image
2. Mask
3. Str (only supports English input)
4. Float

**Model library**
Models can be divided into two types: PREDICT model and EDIT model
PREDICT model list:
1.INVERSE (Subtract mask2 from mask1 or subtract image2 from image1, if mask1 is null it means use a mask with all pixels to be 1 minus the input mask2)
Input: {Mask[mask1], Mask[mask2], Image[image1], Image[image2]}
Output: {Mask[mask], Image[image]}
Constraint: Only can process one kind input per step, if input masks then the images should be null, and if input images then the masks should be null. If input masks it will only output mask, and if input images it will only output image. The mask1 to be null means use a mask with all pixels to be 1 minus the input mask2, for images it can't, the input images must be the same null or same valid not null.
2.RES (Segmentation by object specified by prompt)
Input: {Image[image], Str[prompt]}
Output: {Mask[mask], Image[image]}
Constraint: The given prompt must be in English and if there are locative words or adjectives, include them. The output image is checkerboard transparency visualization, if the user requests to output the segmentation result, then output this image.
3.SOS (Segmentation of main objects in image)
Input: {Image[image]}
Output: {Mask[mask], Image[image]}
Constraint: Unable to perform segmentation on the specified object, can only segment the most prominent target in the image. The output image is checkerboard transparency visualization, if the user requests to output the segmentation result, then output this image.
4.ADD-PRED (Given a prompt and a mask. If mask=null, predict the most appropriate position to add the target represented by this prompt to the image, If mask!=null, this position must be within the given mask)
Input: {Image[image], Str[prompt], Mask[mask]}
Output: {Mask[mask]}
Constraint: The given prompt must be a complete natural language and if there are locative words or adjectives, include them, such as 'add a black dog on the left'. After the mask prediction is completed, the FLUX model needs to be used to complete the editing.
5.CMI-PRED (Describe the image in English, this description is applied to the FLUX model, so that the generated image is inspired by the original image. If it is an image expansion task, the input ratio needs to be given so that the output image and the output mask is output and applied to the FLUX model. Otherwise, the input ratio=null and the output mask=null, the output image=null)
Input: {Image[image], Float[left_ratio], Float[right_ratio], Float[top_ratio], Float[bottom_ratio]}
Output: {Str[caption], Image[image], Mask[mask]}
Constraint: Notice all ratio should be the same null or all no null, can't just one or two to be null. If it is an image expansion task, the output image and the output mask need to be applied to the FLUX model at the same time, don't just use the output mask alone.
6.BBOX (Given a mask, output the bounding box mask of it)
Input: {Mask[mask]}
Output: {Mask[mask]}
Constraint: None

EDIT model list:
1.FASTINPAINT (For quick inpaint and the score of the inpaint effect will be output)
Input: {Image[image], Mask[mask]}
Output: {Image[image], Float[score]}
Constraint: The inpaint effect is poor, it is generally as a pre-inpaint image, if the user is in a hurry, you can also use it directly as the result.
2.FLUX-FILL (Generated in the mask area according to the specified prompt, don't use to replace the color or material of an object)
Input: {Image[image], Mask[mask], Str[prompt], Image[preimage]}
Output: {Image[image]}
Constraint: The input mask must not be None, If the model's input mask is the output mask from CMI-PRED model (like step3[mask]), it's input image must be the output image from CMI-PRED model (step3[image], not step1[image] or step2[image]) too. The input preimage is optional, you can use the original image or reference image or set preimage=null. The model can only be generated according to prompt, if the preimage is a reference image, the input prompt should describe the reference image in detail
3.FLUX-RCM (Replace the color or material of an object)
Input: {Image[image], Mask[mask], Str[prompt]}
Output: {Image[image]}
Constraint: Change the color or material of a specific object according to the input prompt. 
4.FLUX-INPAINT (Fill background in mask area, generate reference the input preimage and the score)
Input: {Image[image], Mask[mask], Image[preimage], Float[score]}
Output: {Image[image]}
Constraint: Cannot be generated according to prompt, can only be used to remove related tasks. The input preimage and score is mandatory, you can use the pre-inpaint image and score from FASTINPAINT model.
5.FLUX-CBG (Can only be used to change the existing background into a new scenery or attraction)
Input: {Image[image], Mask[mask], Str[prompt]}
Output: {Image[image]}
Constraint: The given prompt must be 'change the background to XXX', XXX must be a specific scene, such as 'beach', there must be a previous segmentation model (If explicitly specifying to replace the background of a designated object, use RES model, otherwise, use SOS model) + MASK-INVERSE model to predict the mask. 
6.FLUX-STYLE (Convert the style of the input image or a specific object in the image, you must give an input style, such as 'anime style')
Input: {Image[image], Mask[mask], Str[prompt], Str[style]}
Output: {Image[image]}
Constraint: The given prompt can only be obtained using CMI-PRED model. The default value of input mask=null means whole image style transfer. You can also specify a mask, which means partial style transfer.
7.COMPOSE (Compose two input masks or images, if both have values at same pixels, the second input will cover the first)
Input: {Mask[mask1], Mask[mask2], Image[image1], Image[image2]}
Output: {Mask[mask], Image[image]}
Constraint: Only can process one kind input per step, if input masks then the images should be null, and if input images then the masks should be null. If input masks it will only output mask, and if input images it will only output image.
8.RESIZE (Resize the width and height of the valid part of input mask or image to the given ratio times original width and height)
Input: {Mask[mask], Image[image], Float[ratio]}
Output: {Mask[mask], Image[image]}
Constraint: Input mask and image must have one to be null, only can process one kind input per step. If process mask, only output resized mask, and if process image, only output resized image correspondingly.
9.FLUX-ENV (Replace the environment of an object, like the weather, the climate, or the times of day.)
Input: {Image[image], Str[prompt]}
Output: {Image[image]}
Constraint: Don't use any PREDICT model in advance, change the environment of the scene according to the input prompt. Such as if you want to change the weather to be rainny day, prompt='change the weather to be rainny'.
10.FLUX-POSE (Change the object's posture, expression, etc.)
Input: {Image[image], Str[prompt]}
Output: {Image[image]}
Constraint: The input prompt must provide a detailed description of the external characteristics of the modification target, such as gender, clothing, accessories, etc and don't use any PREDICT model in advance.

**Actual example1:**
User instruction: First add a cat, then expand the image by 2 times
Expected output:
{
  "process": "First add a cat, then expand the image by 2 times",
  "pipeline": [
    {
      "step": 1,
      "model": "ADD-PRED",
      "input": {
        "image": "init[image]",
        "prompt": "cat",
        "mask": null,
      },
      "output": {
        "mask": "step1[mask]"
      }
    },
    {
      "step": 2,
      "model": "FLUX-FILL",
      "input": {
        "image": "init[image]",
        "mask": "step1[mask]",
        "prompt": "cat",
        "preimage": null
      }
      "output": {
        "mask": "step2[image]"
      }
    },
    {
      "step": 3,
      "model": "CMI-PRED",
      "input": {
        "image": "step2[image]",
        "ratio": 2.0
      },
      "output": {
        "caption": "step3[caption]",
        "image": "step3[image]",
        "mask": "step3[mask]"
      }
    },
    {
      "step": 4,
      "model": "FLUX-FILL",
      "input": {
        "image": "step3[image]", 
        "mask": "step3[mask]",
        "prompt": "step3[caption]",
        "preimage": "step2[image]"
      },
      "output": {
        "image": "step4[image]",
      }
    },
    {
      "result": "[step4[image]]"
    }
  ]
}

**Actual example2:**
User instruction: Output the segmentation result of the dog and eliminate the dog
Expected output:
{
  "process": "Output the segmentation result of the dog and eliminate the dog",
  "pipeline": [
    {
      "step": 1,
      "model": "RES",
      "input": {
        "image": "init[image]",
        "prompt": "dog"  
      },
      "output": {
        "mask": "step1[mask]",
        "image": "step1[image]"
      }
    },
    {
      "step": 2,
      "model": "FASTINPAINT",
      "input": {
        "image": "init[image]",
        "mask": "step1[mask]" 
      },
      "output": {
        "image": "step2[image]",
        "score": "step2[score]"
      }
    },
    {
      "step": 3,
      "model": "FLUX-INPAINT",
      "input": {
        "image": "init[image]", 
        "mask": "step1[mask]", 
        "preimage": "step2[image]",
        "score": "step2[score]"
      },
      "output": {
        "image": "step3[image]"
      }
    },
    {
      "result": "[step1[image], step3[image]]"
    }
  ]
}

**Actual example3:**
User instruction: Replace the car with a dog
Expected output:
{
  "process": "Replace the car with a dog",
  "pipeline": [
    {
      "step": 1,
      "model": "RES",
      "input": {
        "image": "init[image]",
        "prompt": "car"  
      },
      "output": {
        "mask": "step1[mask]",
        "image": "step1[image]"
      }
    },
    {
      "step": 2,
      "model": "ADD-PRED",
      "input": {
        "image": "init[image]",
        "prompt": "dog",
        "mask": "step1[mask]",
      },
      "output": {
        "mask": "step2[mask]"
      }
    },
    {
      "step": 3,
      "model": "FASTINPAINT",
      "input": {
        "image": "init[image]",
        "mask": "step1[mask]" 
      },
      "output": {
        "image": "step3[image]",
        "score": "step3[score]"
      }
    },
    {
      "step": 4,
      "model": "FLUX-INPAINT",
      "input": {
        "image": "init[image]", 
        "mask": "step1[mask]",  
        "preimage": "step3[image]",
        "score": "step3[score]"
      },
      "output": {
        "image": "step4[image]"
      }
    },
    {
      "step": 5,
      "model": "FLUX-FILL",
      "input": {
        "image": "step4[image]", 
        "mask": "step2[mask]",
        "prompt": "dog",
        "preimage": null
      },
      "output": {
        "step5[image]"
      }
    },
    {
      "result": "[step5[image]]"
    }
  ]
}

Now, I give you the image and the user instruction: "Your Instruction", please output the multi-model collaboration process json.
\end{lstlisting}

\section{Prompt Example for Producing Ground Truth Data}

Listing~\ref{lst:produce-prompt} provides an exemplar prompt specifically used for generating ground truth training data for the environment change task.

\begin{lstlisting}[style=promptstyle,caption={Prompt Example for Generating Ground Truth Data for the Environment Change Task},label={lst:produce-prompt}]
**System Role**
You are an AI image processing engine scheduler, responsible for converting the natural language instructions provided by the user into executable multi-model collaboration process json. All inputs must come from the initial parameters or the output of the previous model or your own language decomposition and translation.

**Processing rules**
1. Input traceability principle
- Each model parameter can only be:
a. Initial image, it can be an image list composed of many images
b. Text in user instructions
c. Output of previous steps
d. The result of your own language analysis and translation of the text in user instructions

2. Process generation steps
a. Extract the operation object and action from the user instruction and analyze it in combination with the image content
b. After analysis, select the corresponding model for each operation, ensure that the model you choose and the parameters you input meet the requirements of the model
c. Establish a cross-model data dependency chain, make sure the output of each step is used in subsequent processes, otherwise this step is redundant.

**Input and output types**
There are only four types of input and output
1. Image
2. Mask
3. Str (only supports English input)
4. Float

**Model library**
Models can be divided into two types: PREDICT model and EDIT model
PREDICT model list:
None

EDIT model list:
1.FLUX-ENV (Replace the environment of an object, like the weather, the climate, or the times of day)
Input: {Image[image], Str[prompt]}
Output: {Image[image]}
Constraint: Don't use any PREDICT model in advance, change the environment of the scene according to the input prompt. Such as if you want to change the weather to be rainny day, prompt='change the weather to be rainny'.

**Actual example**
Describe: 
1. For tasks that change the environment, you need to consider how to write the prompt. The picture is given, and the prompt should describe the change you want to make.
Example1:
User instruction: Turn the weather into a sunny day with clear blue sky.
Expected output:
{
  "process": "Turn the weather into a sunny day with clear blue sky.",
  "pipeline": [
    {
      "step": 1,
      "model": "FLUX-ENV",
      "input": {
        "image": "init[image]",
        "prompt": "Turn the weather into a sunny day with clear blue sky."
      },
      "output": {
        "image": "step1[image]"
      }
    },
    {
      "result": "[step1[image]]"
    }
  ]
}
Example2:
User instruction: Turn to what the scene looks like in the evening.
Expected output:
{
  "process": "Turn to what the scene looks like in the evening.",
  "pipeline": [
    {
      "step": 1,
      "model": "FLUX-ENV",
      "input": {
        "image": "init[image]",
        "prompt": "Turn to what the scene looks like in the evening."
      },
      "output": {
        "image": "step1[image]"
      }
    },
    {
      "result": "[step1[image]]"
    }
  ]
}

Now, I give you the image and the user instruction: "<INSTRUCTION_TO_REPLACE>"
Please output the thinking process in <think> </think>. During the thinking process
1. Don't mention that you have seen the Actual example. 
2. Carefully think why this collaboration process is used (you can find hints in the 'Describe' section of the Actual example).
Then output the multi-model collaboration process json in <answer> </answer>.
\end{lstlisting}

\end{document}